\PassOptionsToPackage{table}{xcolor}
\RequirePackage{fix-cm}
\documentclass[11pt]{article}
\usepackage[preprint]{acl}
\usepackage{times}
\usepackage{latexsym}
\usepackage[T1]{fontenc}
\usepackage[utf8]{inputenc}
\usepackage{microtype}
\IfFileExists{inconsolata.sty}{\usepackage{inconsolata}}{}
\usepackage{graphicx}
\usepackage{booktabs}
\usepackage{enumitem}
\usepackage{ddrg_visual}
\usepackage{algorithm}
\usepackage{algpseudocode}
\usepackage{amsmath}
\usepackage{amssymb}
\usepackage{tikz}
\usepackage{needspace}
\usetikzlibrary{arrows.meta,positioning,calc}

\hypersetup{
  pdftitle={Beyond Answer Voting: Fine-Grained Correction of LLM Reasoning through Graph Alignment across Sampled Traces},
  pdfauthor={Haotian Liu, Yingjian Chen, Xiaohang Xu, Lang Huang, Yusuke Iwasawa, Yutaka Matsuo, Irene Li},
  breaklinks=true
}

\newcommand{\method}{DDRG}

\title{Beyond Answer Voting: Fine-Grained Correction of LLM Reasoning through Graph Alignment across Sampled Traces}

\author{
\textbf{Haotian Liu}$^{1}$ \quad
\textbf{Yingjian Chen}$^{2}$ \quad
\textbf{Xiaohang Xu}$^{2}$ \quad
\textbf{Lang Huang}$^{3}$ \\
\textbf{Yusuke Iwasawa}$^{2}$ \quad
\textbf{Yutaka Matsuo}$^{2}$ \quad
\textbf{Irene Li}$^{2}$ \\[0.5em]
{\normalfont\small $^{1}$School of Informatics, Xiamen University} \\
{\normalfont\small $^{2}$The University of Tokyo \quad
$^{3}$National Institute of Informatics (NII)} \\[0.4em]
{\normalfont\small
\texttt{haotianliu.me@gmail.com} \quad
\texttt{irene.li@weblab.t.u-tokyo.ac.jp}}
}

\begin{document}
\maketitle

\begin{abstract}
% Sampling multiple reasoning traces can improve language-model answers, but final-answer voting leaves intermediate disagreements unresolved. We introduce Disagreement-Driven Verification Guided by Reasoning Graphs (DDRG), which turns these disagreements into evidence for selective correction. DDRG extracts source-grounded graphs from a fixed bank of ordinary reasoning traces, aligns competing claims, and asks a separate verifier to assess them. Evidence against the voted answer's supporting reasoning guides one new response; otherwise, the original vote is retained. Evaluations with two solvers across mathematical, logical, and medical question answering show net improvements over self-consistency on the identical candidate banks without task-specific verifier fine-tuning. Paired analysis distinguishes recovered candidate answers, newly generated correct answers, and harmful changes. The results suggest that reasoning diversity can support correction through its intermediate claims, rather than only through final-answer frequency.
Large language models increasingly tackle tasks that require multi-step reasoning, where errors in intermediate claims can affect the final answer. Self-consistency (SC) improves reasoning by voting over sampled final answers, but its vote leaves intermediate disagreements unresolved. To address this problem, we introduce \method{} (Disagreement-Driven Verification Guided by Reasoning Graphs), a framework for selective correction from a frozen bank of reasoning traces. Our insight is that graph alignment can expose claim-level disagreements and provide correction evidence without expanding the sampled trace bank. \method{} builds source-grounded claim graphs and sends localized conflicts to an answer-blind verifier. A verifier-supported refutation that reaches the majority answer's supporting path triggers one focused repair; otherwise, \method{} returns the exact SC vote. Experimental results across five benchmarks show that native zero-shot \method{} improves macro-average accuracy over direct answering by 8.65 and 15.83 percentage points for Qwen2.5-7B and Llama3.1-8B. It also exceeds CoT, the evaluated GoT and GoV question-answering adaptations, and SC on this macro metric. \method{} links local verification to selective correction without enlarging the initial candidate bank\footnote{Our anonymous codebase is available at \url{https://anonymous.4open.science/r/DDRG-3D31/}}.

\end{abstract}

\section{Introduction}
\label{sec:intro}

% Large language models (LLMs) increasingly solve problems that require linked
% inferences. Chain-of-thought (CoT) exposes intermediate reasoning
% \citep{wei2022chain}, and sampling multiple traces turns reasoning diversity
% into a test-time resource \citep{wang2023selfconsistency}. Self-consistency
% (SC) selects the most frequent final answer, yet an endpoint vote cannot show
% which claim caused a disagreement or whether a minority trace contains a
% repair. A useful correction mechanism should treat the sampled traces as
% evidence: localize a conflicting claim, check it independently, and determine
% whether it can change the answer.
Large language models (LLMs) increasingly tackle problems that require multi-step reasoning, including mathematical problem solving, logical inference, and medical reasoning~\citep{lewkowycz2022solving, singhal2023large, pan2023logic}. These tasks require models to derive intermediate claims and use them to reach a final answer. Chain-of-thought (CoT) prompting makes these steps explicit \citep{wei2022chain}, but the resulting traces can still contain errors. An incorrect intermediate claim can affect later steps and lead to a wrong answer, creating a need for methods that can identify and correct reasoning errors.

\begin{figure}[t!]
\begin{nolinenumbers}
\centering
\includegraphics[width=0.48\textwidth]{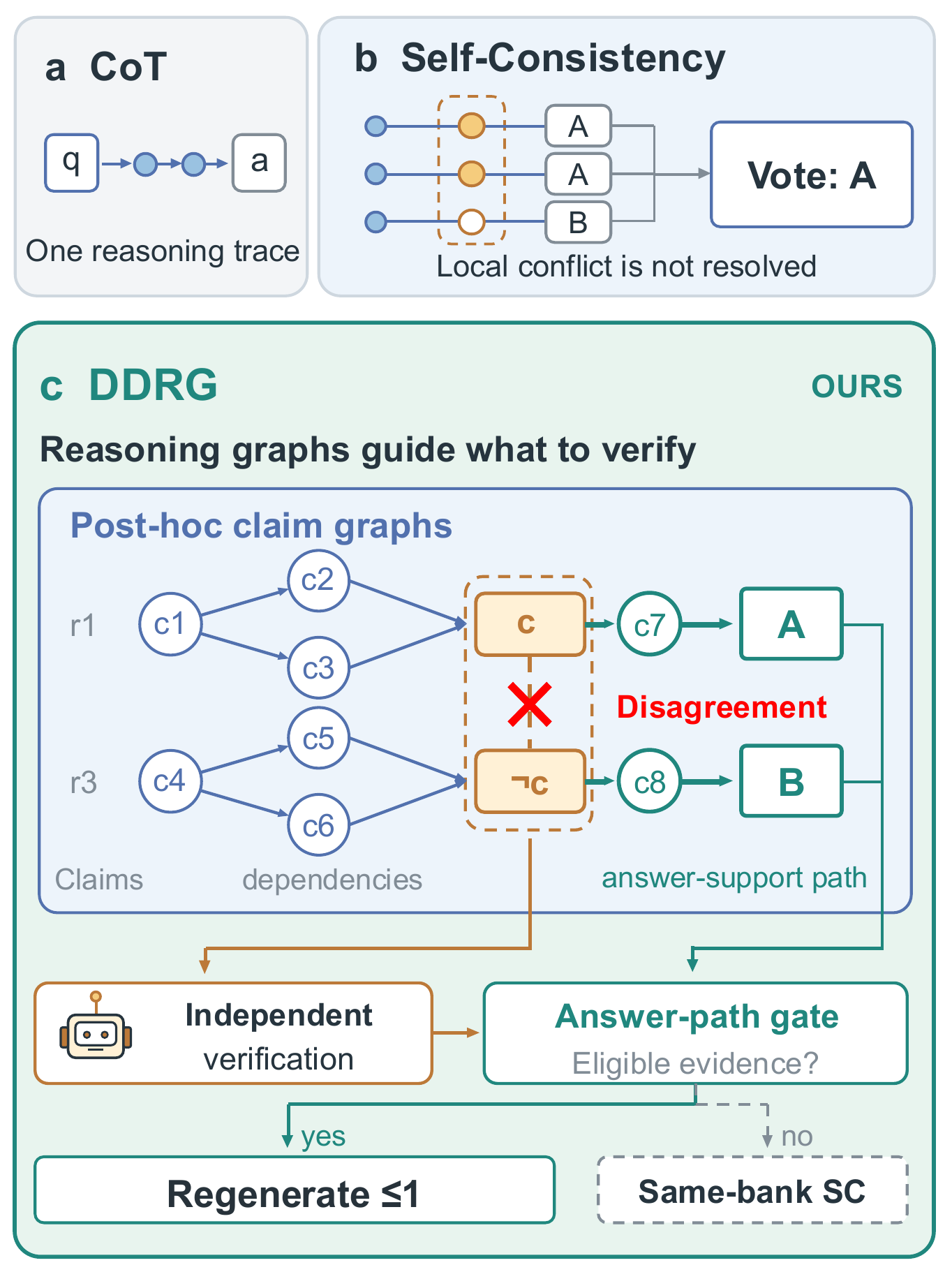}
\caption{Comparison of CoT, self-consistency (SC), and DDRG. CoT uses one trace; SC votes over final answers. DDRG verifies aligned local disagreements and uses answer-relevant evidence for one regeneration, retaining SC when correction is unwarranted.}
\label{fig:motivation}
\end{nolinenumbers}
\end{figure}

% Graph of Thoughts and Graph of Verification organize or check intermediate
% reasoning \citep{besta2024graph,fang2025gov}, motivating structured test-time
% computation. Their endpoint still needs a mechanism that preserves the local
% claim, its provenance, and its dependency path. DDRG keeps the frozen CoT
% bank, builds a source-grounded post-hoc graph for each trace, aligns claims
% about the same local object, and sends the conflict to an answer-blind
% verifier. A verifier-supported refutation that reaches the majority answer path
% triggers one focused response from the solver; otherwise the exact SC vote is
% returned. The graph turns trace diversity into a bounded correction decision.

% Across five mathematical, logical, and medical cohorts, native zero-shot DDRG
% improves the identical same-bank SC reference by 0.65 and 0.41 macro points
% for the two solver panels. Most successful Qwen rescues recover a correct
% minority answer already present in the bank, and matched controls support a
% graph-routing signal on the registered mechanism subset. These results show
% that source-grounded graphs can make sampled reasoning diversity actionable for
% compact solver models.

Existing methods improve reasoning by sampling multiple traces, organizing intermediate steps, or checking their validity. Graph of Thoughts and Graph of Verification explore structured reasoning and verification \citep{besta2024graph,fang2025gov}. Self-consistency (SC) remains a simple and effective baseline: it samples several CoT traces and selects the most frequent final answer \citep{wang2023selfconsistency} (Figure~\ref{fig:motivation}(b)). This answer-level vote leaves intermediate disagreements unresolved. It does not reveal which claims cause traces to disagree or whether a minority trace contains a valid correction.

However, the intermediate claims in SC's sampled traces can be used more effectively. Our insight is that graph representations make it natural to align corresponding claims across traces and pinpoint local disagreements. This allows fine-grained use of existing samples without expanding the CoT bank. Motivated by this, we introduce \textbf{DDRG} (\textbf{D}isagreement-\textbf{D}riven Verification Guided by \textbf{R}easoning \textbf{G}raphs), shown in Figure~\ref{fig:motivation}(c). DDRG builds a source-grounded post-hoc graph for each trace in a frozen CoT bank, preserving quoted claims, their provenance, and their dependencies. It aligns claims about the same local object and sends a localized conflict to an independent, answer-blind verifier. When a verifier-supported refutation reaches the majority answer's supporting path, DDRG requests one focused repair. Otherwise, it returns the exact SC vote from the same bank. We evaluate native zero-shot DDRG on five cohorts covering mathematical, logical, and medical reasoning. DDRG achieves higher macro-average accuracy than direct answering, CoT, the evaluated GoT and GoV adaptations, and SC.

Our contributions are:
\begin{itemize}[noitemsep, topsep=1pt]
    \item \textbf{Introducing graph-based fine-grained correction.} We propose DDRG, a source-grounded graph framework that turns disagreements among sampled reasoning traces into evidence for correcting majority answers. Its core insight is that graph alignment exposes claim-level disagreements without expanding the CoT bank.
    \item \textbf{Providing a reusable protocol for selective repair.} We introduce a correction protocol across heterogeneous reasoning tasks that supports answer-blind verification, at most one focused repair, and exact SC fallback.
    \item \textbf{Empirical gains and minority-answer recovery.} We evaluate two locally deployed solvers on MedQA, GSM8K, ReClor, LogiQA2, and FOLIO. Macro-average gains over direct answering are 8.65 and 15.83 percentage points for Qwen2.5-7B and Llama3.1-8B, respectively. Our analysis shows recovery of correct candidate answers when SC is wrong.
\end{itemize}

\section{Related Work}
\label{sec:related}
\paragraph{Reasoning diversity at test time.}
Test-time methods improve reasoning by exploring alternative paths. SC aggregates answers from sampled CoT traces, while Tree of Thoughts searches reasoning branches \citep{wei2022chain,wang2023selfconsistency,yao2023tree}. Adaptive approaches further allocate computation according to problem difficulty \citep{snell2025scaling}. In contrast to existing methods, DDRG compares intermediate claims across existing traces to locate disagreements without expanding the sampled bank.

\paragraph{Graph-structured reasoning.}
Graph-based methods represent reasoning steps and their relationships explicitly. Graph of Thoughts combines and refines thoughts, while Graph of Verification checks reasoning units in dependency order \citep{besta2024graph,fang2025gov}. Reasoning Consensus aggregates graphs through cross-trace support, and ARGORA models support and attack relations among arguments \citep{parulekar2026reasoning,jin2026argora}. DDRG instead uses source-grounded graphs to align claims across traces and connect local disagreements to the voted answer's supporting path.

\paragraph{Verification and correction.}
Existing methods revise outputs using verbal feedback \citep{madaan2023selfrefine,shinn2023reflexion}, verification questions \citep{weng2023selfverification,dhuliawala2024cove}, or tools and retrieval \citep{gou2024critic,gao2023rarr}. SETS combines sampling, verification, and correction, while SCoRe learns multi-turn correction \citep{ren2025sets,kumar2025score}. Error-localization studies further examine how models identify faulty reasoning steps \citep{tyen2024errors,zheng2025processbench}. In contrast, DDRG derives verification targets from cross-trace disagreements and uses independently verified evidence to guide selective repair.

\begin{figure*}[t!]
\begin{nolinenumbers}
\centering
\includegraphics[width=\textwidth]{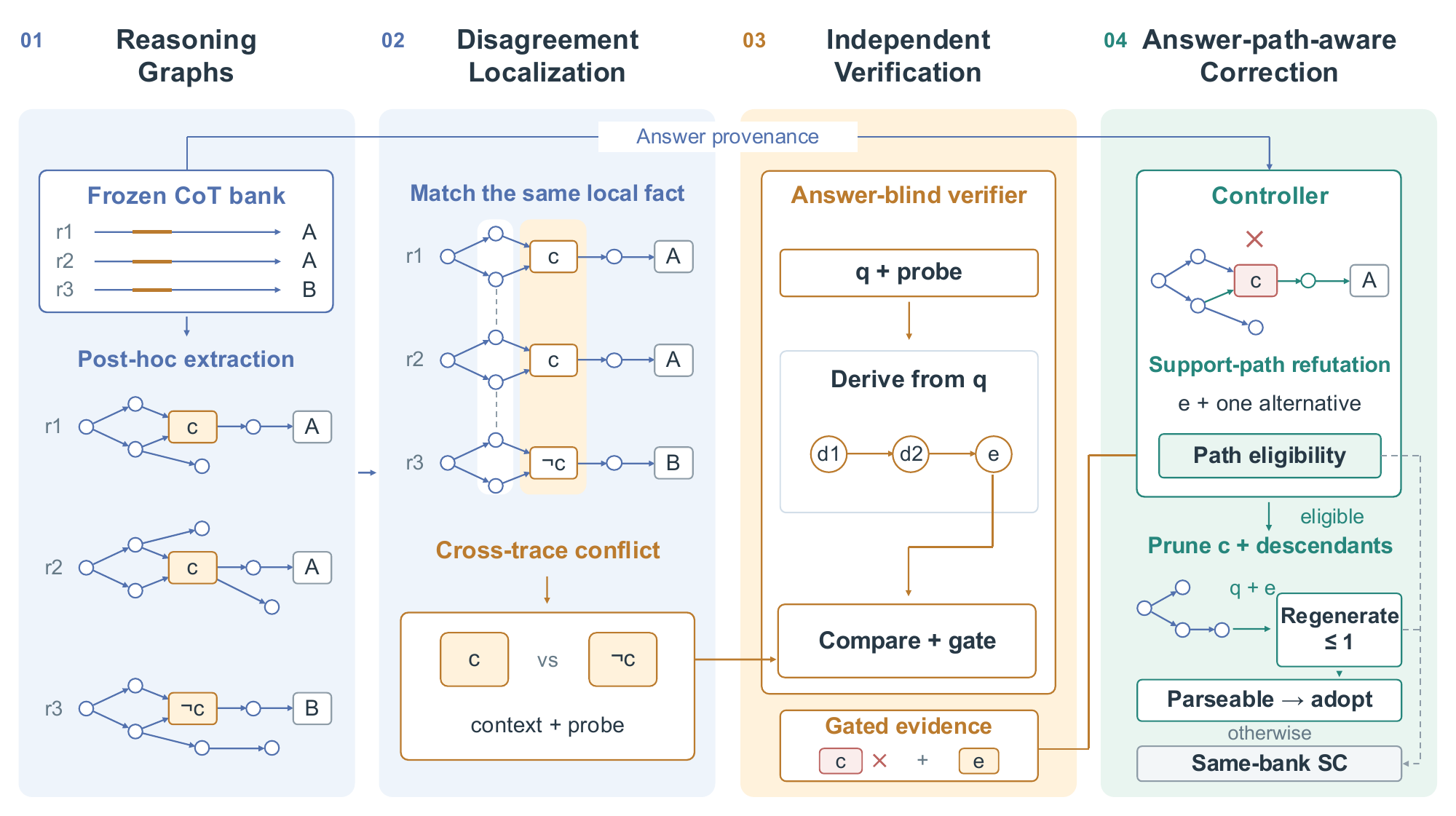}
\caption{DDRG overview. SC and DDRG share a frozen bank. The verifier assesses local claims; the controller tests answer-path relevance and permits one regeneration. Ineligible evidence or an unparseable response returns SC.}
\label{fig:pipeline}
\end{nolinenumbers}
\end{figure*}

\Needspace{8\baselineskip}

\section{Method}
\label{sec:method}
\textbf{DDRG} turns cross-trace disagreements into evidence for selective correction (Figure~\ref{fig:pipeline}). It first extracts source-grounded graphs from a frozen CoT bank, then aligns competing local claims and verifies them independently. The controller uses refutations on the voted answer's support path to permit at most one regeneration; otherwise it retains the exact same-bank SC vote. Algorithm~\ref{alg:ddrg} gives the full procedure.

\Needspace{7\baselineskip}
\subsection{Reasoning Graph Construction}
\label{sec:bank}
For a question $q$, DDRG uses a frozen bank of $K=3$ CoT traces $r_i$ and their parsed answers $a_i$. The parser returns a nonempty answer or $\bot$ on failure. The bank and its default SC answer are
\begin{equation}
\begin{aligned}
\mathcal B(q) &= [(r_i,a_i)]_{i=1}^{K},\\
a_{\mathrm{maj}} &= \operatorname{Vote}(\mathcal B(q)).
\end{aligned}
\label{eq:bank}
\end{equation}
Each parseable answer contributes one vote, including repeated answers. Ties are broken by the earliest bank position. If none parses, Vote returns the original SC fallback.

The solver $S$ converts each trace into a reasoning graph $G_i=(V_i,E_i)$. Each node stores an exact claim quotation and its location in the source trace. Edges record the support relations inferred by the solver. These graphs allow DDRG to compare claims across traces and trace their connections to the final answer. Graphs that fail the source or structure checks are skipped. Let $\mathcal G(q)$ denote the remaining indexed graphs. Graph failures leave the frozen bank and its SC vote unchanged.

\subsection{Disagreement Localization}
\label{sec:localize}
DDRG aligns claims about the same quantity, entity, relation, or proposition across traces. It collects conflicting claims into an ordered sequence of disputes:
\begin{equation}
\begin{aligned}
\mathcal D(q) &= (d_1,\ldots,d_m),\\
d_j &= (c_{\ell,j},c_{r,j},C_j,p_j).
\end{aligned}
\label{eq:disputes}
\end{equation}
Each dispute contains two competing claims, nearby context $C_j$, and a probe question $p_j$. Both claims must be exact quotations linked to their source graph nodes. Node references include the trace index and node identifier.

The answers associated with the source traces may be used to prioritize disputes. Final-answer nodes are excluded from verification targets. If no dispute passes the source and node-reference checks, DDRG returns $a_{\mathrm{maj}}$.

\subsection{Independent Verification}
\label{sec:verify}
A verifier $V$ checks at most the first three valid disputes in their recorded order. For each dispute, it receives $(q,c_\ell,c_r,C,p)$. It first derives a local fact from $q$ and $p$, then compares the conflicting claims against that fact using context $C$. The resulting judgment record $z$ contains a directional verdict, the local evidence $e(z)$, and its derivation $\rho(z)$.

The verifier does not receive candidate final-answer fields, the SC vote, or gold labels. Source quotations may still contain quantities that appear in candidate answers. Let $\hat y(z)$ denote the generated verdict, $y_{\mathrm{native}}(z)$ the highest-scored native class, and $s_{\mathrm{norm}}(z)$ the normalized conditional score. Acceptance requires successful parsing, source grounding, the fixed zero-margin check, and $\hat y(z)=y_{\mathrm{native}}(z)$ with $s_{\mathrm{norm}}(z)\ge 0.5$. Appendix~\ref{sec:decision_details} defines the five-way zero-shot interface and scoring rules. Let $\mathcal J_{\mathrm{acc}}(q)$ collect judgments that pass these checks and support one claim while concretely refuting its counterpart. All other comparisons provide no correction evidence.

\subsection{Answer-Path-Aware Correction}
\label{sec:regenerate}
If no answer is parseable, DDRG returns the baseline SC output. Otherwise, before checking graph usability, the controller fixes the earliest trace with answer $a_{\mathrm{maj}}$ as the representative $i_{\mathrm{rep}}$.

For a usable representative graph and answer node $v_{\mathrm{ans}}$, define the nonterminal ancestor set as
\begin{equation}
\mathcal A_{\mathrm{rep}}(q)=\operatorname{Anc}^{\mathrm{nt}}_{G_{i_{\mathrm{rep}}}}(v_{\mathrm{ans}}).
\label{eq:answer_ancestors}
\end{equation}
Here, $\operatorname{Anc}^{\mathrm{nt}}_G(v)$ contains source-claim nodes with a directed support path of positive length to $v$, excluding question and answer nodes. Set $\mathcal A_{\mathrm{rep}}(q)=\varnothing$ if the graph or answer node is unusable. Accepted evidence is eligible when it refutes a node in $\mathcal A_{\mathrm{rep}}(q)$, provides a nonempty independent derivation, and associates its supported claims with one common candidate-answer provenance value other than $a_{\mathrm{maj}}$. Decisive support for $a_{\mathrm{maj}}$ blocks correction. Let $\mathcal E(q)$ collect the evidence passing these checks.

For eligible evidence, the controller removes refuted nodes and their dependent descendants, retaining the remaining claims as unrefuted context $C^\star$. The solver generates one response $r^\star=S(q,C^\star,\mathcal E(q))$. Set $a^\star=\operatorname{Parse}(r^\star)$ when regeneration occurs and $a^\star=\bot$ otherwise. The final answer is
\begin{equation}
\hat a(q)=
\begin{cases}
a^\star, & a^\star\ne\bot,\\
a_{\mathrm{maj}}, & a^\star=\bot.
\end{cases}
\label{eq:adoption}
\end{equation}
Any parseable regenerated answer is adopted, including answers outside the bank, without a second correctness check. Otherwise, DDRG returns the exact SC vote. Service retries do not count as additional logical regenerations.

% \begin{algorithm}[t]
% \caption{DDRG: selective correction from a frozen reasoning bank}
% \label{alg:ddrg}
% \begin{algorithmic}[1]
% \Require Question $q$; solver $S$; local verifier $V$; $K=3$
% \State $\mathcal B\gets \operatorname{SampleAndFreeze}(S,q,K)$
% \State $a_{\mathrm{maj}}\gets \operatorname{Vote}(\mathcal B)$
% \State $\mathcal G\gets \operatorname{SourceGraphs}(S,\mathcal B)$
% \State $\mathcal D\gets \operatorname{AlignAndLocalize}(S,\mathcal G,q)$
% \State $\mathcal J\gets \operatorname{Verify}(V,\operatorname{FirstValid}_3(\mathcal D),q)$
% \State $\mathcal E\gets \operatorname{Eligible}(\mathcal J,\mathcal G,\mathcal B,a_{\mathrm{maj}})$
% \If{$\mathcal E=\varnothing$}
%   \State \Return $a_{\mathrm{maj}}$
% \EndIf
% \State $C^\star\gets \operatorname{RetainUnaffected}(\mathcal G,\mathcal E)$
% \State $a^\star\gets\parse(S(q,C^\star,\mathcal E))$ \Comment{one regeneration}
% \State \Return $a^\star$ \textbf{if} $a^\star\ne\bot$, \textbf{else} $a_{\mathrm{maj}}$
% \end{algorithmic}
% \end{algorithm}
\begin{algorithm}[t]
\caption{DDRG: selective correction from a frozen CoT bank}
\label{alg:ddrg}
\begin{algorithmic}[1]
\Require Question $q$; solver $S$; verifier $V$; $K=3$
\State $\mathcal B\gets\operatorname{SampleAndFreeze}(S,q,K)$
\State $a_{\mathrm{maj}}\gets\operatorname{Vote}(\mathcal B)$
\State $\mathcal I\gets\{i\in\{1,\ldots,K\}:a_i\ne\bot\}$
\If{$\mathcal I=\varnothing$}
    \State \Return $a_{\mathrm{maj}}$
\EndIf
\State $i_{\mathrm{rep}}\gets\min\{i\in\mathcal I:a_i=a_{\mathrm{maj}}\}$
\State $\mathcal G\gets\operatorname{SourceGraphs}(S,\mathcal B)$
\State $\mathcal D\gets\operatorname{AlignAndLocalize}(S,\mathcal G,q)$
\State $\mathcal D_{\mathrm{ver}}\gets\operatorname{FirstValid}_3(\mathcal D)$
\State $\mathcal J_{\mathrm{acc}}\gets\operatorname{VerifyAndGate}(V,q,\mathcal D_{\mathrm{ver}})$
\State $\mathcal E\gets\operatorname{RouteEvidence}(\mathcal J_{\mathrm{acc}},\mathcal G,\mathcal B,i_{\mathrm{rep}})$
\If{$\mathcal E=\varnothing$}
    \State \Return $a_{\mathrm{maj}}$
\EndIf
\State $C^\star\gets\operatorname{RetainUnaffected}(\mathcal G,\mathcal E)$
\State $r^\star\gets S(q,C^\star,\mathcal E)$ \Comment{one regeneration}
\State $a^\star\gets\operatorname{Parse}(r^\star)$
\If{$a^\star\ne\bot$}
    \State \Return $a^\star$
\Else
    \State \Return $a_{\mathrm{maj}}$
\EndIf
\end{algorithmic}
\end{algorithm}
\section{Experiments}
\label{sec:experiments}

\begin{table*}[t]
\begin{nolinenumbers}
\centering
% Native Llama manifest SHA256 b91163380175b0da15590d5396e1e8798cfded5cb6b77fdef8c1e522628b0cdc
% Qwen and baseline numbers retain their frozen manifests.
% Presentation-only change. Every data cell is bound to the frozen source.
% Requires \usepackage{ddrg_visual}; colors identify roles, not rankings.
\begingroup
\normalsize
\setlength{\tabcolsep}{3pt}
\renewcommand{\arraystretch}{1.16}
\resizebox{0.88\linewidth}{!}{%
\begin{tabularx}{\linewidth}{@{}>{\raggedright\arraybackslash}p{0.30\linewidth}*{6}{>{\centering\arraybackslash}X}@{}}
\toprule
\textbf{Method} & \textbf{MedQA} & \textbf{GSM8K} & \textbf{ReClor} & \textbf{LogiQA2} & \textbf{FOLIO} & \shortstack{\textbf{Macro}\\\textbf{Avg.}} \\
\midrule
\multicolumn{7}{l}{\textbf{Qwen2.5-7B}} \\
Direct & 60.88 & 70.36 & 70.00 & 63.80 & 55.39 & 64.09 \\
CoT \citep{wei2022chain} & 62.77 & 89.01 & 72.00 & 65.39 & \textbf{68.63} & 71.56 \\
GoT (QA) \citep{besta2024graph} & 63.16 & 90.90 & 70.40 & 64.89 & 59.31 & 69.73 \\
GoV (QA) \citep{fang2025gov} & 60.64 & 90.22 & 73.40 & 63.42 & 58.33 & 69.20 \\
\addlinespace[2pt]
\rowcolor{DDRGTealLight}
\DDRGMethod{DDRG} & \textbf{64.02} & \textbf{92.27} & \textbf{74.40} & \textbf{67.30} & 65.69 & \textbf{72.74} \\
\midrule
\multicolumn{7}{l}{\textbf{Llama3.1-8B}} \\
Direct & 53.10 & 51.78 & 56.20 & 51.53 & 41.18 & 50.76 \\
CoT \citep{wei2022chain} & 68.42 & 84.23 & 61.00 & 53.94 & 54.41 & 64.40 \\
GoT (QA) \citep{besta2024graph} & 68.74 & 84.08 & \textbf{64.80} & \textbf{56.36} & 50.49 & 64.89 \\
GoV (QA) \citep{fang2025gov} & 59.31 & 74.00 & 59.60 & 52.93 & 44.12 & 57.99 \\
\addlinespace[2pt]
\rowcolor{DDRGTealLight}
\DDRGMethod{DDRG} & \textbf{70.07} & \textbf{88.48} & 64.00 & 55.47 & \textbf{54.90} & \textbf{66.58} \\
\bottomrule
\end{tabularx}%
}
\endgroup

\caption{Full-cohort accuracy (\%). Macro averages equally weight the five benchmarks. Bold marks the best result among the five displayed methods within each solver. GoT and GoV are QA adaptations (Section~\ref{subsec:setup}); SC and same-bank gains are reported separately in Appendix~\ref{sec:complete_comparison}.}
\label{tab:main_results}
\end{nolinenumbers}
\end{table*}
We evaluate the complete procedure against the same-bank vote, then use component controls and saved cases to examine how local evidence changes an answer.

\subsection{Experimental Setup}
\label{subsec:setup}
\paragraph{Benchmarks.}
We evaluate Qwen2.5-7B-Instruct and Llama3.1-8B-Instruct on MedQA test, GSM8K test, ReClor validation, LogiQA2 test, and FOLIO public v0.0 validation \citep{jin2021medqa,cobbe2021training,reclor2020,logiqa2023,han2024folio}. Their respective sizes are 1,273, 1,319, 500, 1,572, and 204: 4,868 labeled questions per solver. An additional 1,000 ReClor hidden-test predictions are not scored. The same correction procedure is used across these mathematical, logical, and medical settings.

\paragraph{Baselines.}
Baselines are Direct, independent single-response CoT, SC@3, and question-answering adaptations of Graph of Thoughts (GoT) and Graph of Verification (GoV). Direct and CoT use deterministic decoding. SC samples three responses at temperature 0.9 and top-$p$ 0.95 with a 4,096-token budget; DDRG reuses these exact responses and their parsed answers. GoT generates three thoughts, then scores, aggregates, and refines them in six calls. GoV checks up to 12 paragraph blocks with preceding verified context, stops at the first refutation, and permits one repair. It does not execute a separate graph-extraction stage. These adaptations, their prompts, and their stage budgets are specified in Appendix~\ref{sec:full_prompts}.

\paragraph{Implementation Details.}
Both DDRG solvers use official Qwen3-30B-A3B-Instruct-2507 without project fine-tuning. Native scoring, generated/scored agreement, and source-grounding checks govern evidence acceptance. The verifier context is 8K. Llama retains an 8K solver context except for four preregistered whole-block 32K recoveries; banks, prompts, decoding, and verifier gates remain fixed. Appendix~\ref{sec:complete_setup} records revisions and recovery identities. Gold labels enter offline scoring only; invalid predictions remain in the denominator.

\paragraph{Metrics.}
Accuracy is reported per benchmark, with an equal-benchmark macro average. Paired wrong-to-right (W$\!\to\!$R) and right-to-wrong (R$\!\to\!$W) transitions quantify corrections and introduced errors relative to SC. Their difference is a pooled count, not a macro percentage. For Qwen, we also report the existing exact McNemar test and a benchmark-stratified paired bootstrap interval, conditional on the frozen generations (Appendix~\ref{sec:statistical_details}).

\subsection{Main Results}
\label{subsec:main_results}
Table~\ref{tab:main_results} compares DDRG with direct answering, independent single-response CoT, and two graph-based QA adaptations. DDRG achieves the highest \emph{benchmark-macro} accuracy among the evaluated methods for both solvers: 72.74\% with Qwen and 66.58\% with Llama. Compared with independent CoT, these are gains of 1.18 and 2.18 percentage points. DDRG improves on CoT in nine of ten solver--benchmark settings; the exception is Qwen FOLIO. The largest CoT-relative gains occur on GSM8K (+3.26 and +4.25 points for Qwen and Llama), indicating that the aggregate effect is not confined to a single solver. These independent-CoT comparisons characterize performance, but do not isolate the effect of correction on an identical set of sampled traces.

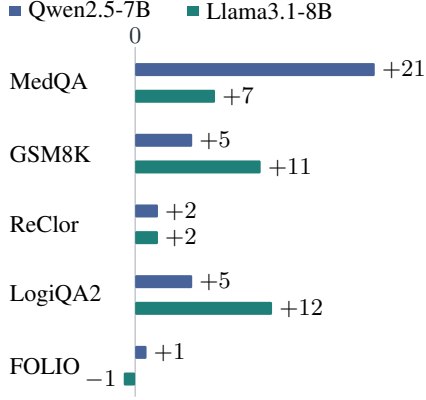
\begin{figure}[t]
\begin{nolinenumbers}
\centering
% Same-bank paired net correction counts.
% Qwen [21,5,2,5,1], Llama [7,11,2,12,-1].
% Totals: Qwen +34, Llama +31.
% Uses official ACL template fonts.

\begin{tikzpicture}[
    x=0.82cm,
    y=0.82cm,
    baseline=(current bounding box.center)
  ]

  % Consistent smaller typography
  \tikzset{
    netlabel/.style={
      font=\normalfont\footnotesize,
      inner sep=0pt
    }
  }

  % ------------------------------------------------
  % Legend
  % ------------------------------------------------
  \fill[DDRGBlue!88!black]
    (0.08,6.38) rectangle (0.27,6.53);

  \node[netlabel,anchor=west]
    at (0.38,6.455) {Qwen2.5-7B};

  \fill[DDRGTeal!95!black]
    (2.95,6.38) rectangle (3.14,6.53);

  \node[netlabel,anchor=west]
    at (3.25,6.455) {Llama3.1-8B};

  % ------------------------------------------------
  % Vertical zero reference
  % ------------------------------------------------
  \draw[DDRGGray!65,line width=.32pt]
    (2.11,0.24) -- (2.11,5.88);

  \node[
    netlabel,
    anchor=south,
    text=DDRGInk
  ] at (2.11,5.93) {$0$};

  % ------------------------------------------------
  % Dataset rows
  % ------------------------------------------------
  \foreach \name/\q/\l/\yy in {
    MedQA/21/7/5.30,
    GSM8K/5/11/4.16,
    ReClor/2/2/3.02,
    LogiQA2/5/12/1.88,
    FOLIO/1/-1/0.74
  }{

    % Dataset name
    \node[netlabel,anchor=west]
      at (0.08,\yy) {\name};

    % Bar endpoints
    \pgfmathsetmacro{\qx}{2.11+0.184*\q}
    \pgfmathsetmacro{\lx}{2.11+0.184*\l}

    % Qwen bar
    \fill[
      DDRGBlue!88!black,
      rounded corners=.5pt
    ]
      (2.11,\yy+0.11)
      rectangle (\qx,\yy+0.32);

    % Llama bar
    \fill[
      DDRGTeal!95!black,
      rounded corners=.5pt
    ]
      ({min(2.11,\lx)},\yy-0.32)
      rectangle ({max(2.11,\lx)},\yy-0.11);

    % Qwen numeric label
    \node[netlabel,anchor=west]
      at (\qx+0.11,\yy+0.215) {$+\q$};

    % Llama numeric label
    \ifnum\l<0
      \node[netlabel,anchor=east]
        at (\lx-0.11,\yy-0.215) {$\l$};
    \else
      \node[netlabel,anchor=west]
        at (\lx+0.11,\yy-0.215) {$+\l$};
    \fi

  }

\end{tikzpicture}
\caption{Net corrections over same-bank SC. In each pair, the upper bar is Qwen and the lower bar is Llama. Net counts subtract harmful from beneficial changes; totals are $+34$ and $+31$. Dataset sizes differ.}
\label{fig:net_corrections}
\end{nolinenumbers}
\end{figure}

The most direct controlled reference is therefore the \emph{same-bank self-consistency} (SC) vote. The three CoT candidates, their order, and the default vote are identical for SC and DDRG; only the subsequent graph-guided verification and correction stages differ. Per-benchmark SC scores appear in Appendix~\ref{sec:complete_comparison}. Across the five benchmarks, DDRG raises SC macro accuracy from 72.09\% to 72.74\% with Qwen and from 66.17\% to 66.58\% with Llama. The recorded outcomes comprise 63 wrong-to-right (W$\!\to\!$R) versus 29 right-to-wrong (R$\!\to\!$W) changes for Qwen, and 38 versus seven for Llama. The resulting net gains, 34 and 31 correct answers, are shown by benchmark in Figure~\ref{fig:net_corrections}. These are pooled counts across questions, whereas the reported macro changes weight datasets equally.

\paragraph{Where the gains occur.}
The positive net corrections are distributed across distinct task types rather than concentrated on the same benchmark for both solvers (Figure~\ref{fig:net_corrections}). Qwen gains 21 answers on medical question answering (MedQA), accounting for 21 of its 34 net corrections; it also gains five each on GSM8K and LogiQA2, two on ReClor, and one on FOLIO. Llama gains most on logical reasoning (LogiQA2, $+12$) and arithmetic word problems (GSM8K, $+11$), followed by MedQA ($+7$) and ReClor ($+2$); FOLIO has one net harmful change ($-1$). Qwen has positive net corrections on all five benchmarks, whereas Llama has positive gains on four and a one-question net loss on FOLIO. Their largest gains occur in different domains.

\paragraph{Counts versus accuracy.}
Because the benchmark sizes differ, the plotted counts should not be interpreted as normalized accuracy improvements. For example, Qwen's 21 net additional correct MedQA answers correspond to $+1.65$ percentage points, whereas Llama's 11 net additional GSM8K correct answers correspond to $+0.83$ points, calculated from the exact counts. The benchmark-macro results in Table~\ref{tab:main_results} and the paired SC accuracies in Appendix~\ref{sec:complete_comparison} therefore remain necessary. Section~\ref{subsec:intervention} separately measures how often the correction mechanism preserves a correct vote or rescues an incorrect one.

\paragraph{What the paired audit adds.}
Unlike independent CoT, same-bank SC holds the initial candidate answers fixed. Paired transitions therefore expose both recovered answers and errors introduced by the complete verification-and-correction procedure. The $+34$ and $+31$ net counts describe the same outcomes as the macro gains, not additional successes. In particular, accepted verification evidence does not guarantee a correct regenerated answer.

\paragraph{Comparison boundaries.}
DDRG does not dominate every individual task: independent CoT leads Qwen FOLIO (68.63\%), GoT leads Llama ReClor and LogiQA2 (64.80\% and 56.36\%), and same-bank SC exceeds DDRG on Llama FOLIO (55.39\% versus 54.90\%). The evidence supports an improvement of the complete post-hoc verification-and-correction procedure under the frozen-bank evaluation, not a uniform or graph-only advantage. Section~\ref{sec:analysis} examines which answer-disagreement patterns and intervention outcomes account for these aggregate differences.

\begin{figure*}[t]
\begin{nolinenumbers}
\centering
\includegraphics[width=0.97\textwidth]
  {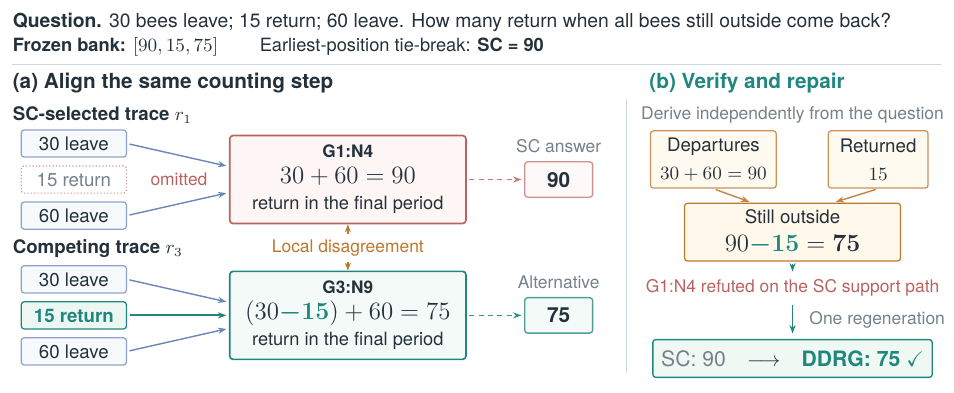}
\caption{DDRG corrects GSM8K \texttt{test:147}
from SC's 90 to 75. Aligned claim fragments expose
the omitted earlier returns; independent verification
and answer-path evidence trigger one regeneration.
The illustrated local dependencies are condensed
(Appendix~\ref{sec:full_cases}).}
\label{fig:case_gsm8k}
\end{nolinenumbers}
\end{figure*}

\subsection{Ablation and Evidence-use Controls}
\label{subsec:ablations}
\paragraph{Graph routing and answer-path restriction.}
The registered subset contains 32 questions from each benchmark. Controls use the same Qwen bank, native verifier, three-probe cap, and at most one regeneration. The graph-free control replaces inferred edges and ancestor routing with fixed flat-source rules. The without-path control retains graphs and verification but removes the majority support-path restriction. Newly triggered regenerations are actually executed.

Full DDRG obtains 118 correct answers, between graph-free (117) and without-path (119), with SC at 116 (Appendix~\ref{sec:mechanism_controls}). Relative to full DDRG, graph-free gains two answers and loses three; without-path gains two and loses one. These small paired differences describe alternative intervention choices rather than establish the necessity of either restriction. Appendix~\ref{sec:mechanism_controls} gives their definitions and the supplementary 49-question non-unanimous slice. Full-cohort disagreement analysis is kept separate in Section~\ref{sec:analysis}.

\paragraph{Selection versus regeneration.}
A retrospective replay keeps the bank, recorded evidence, and eligibility fixed but selects the unique supported candidate instead of generating a response. It reaches 72.42\% Qwen macro accuracy, versus 72.74\% for actual regeneration. Selection produces 85 W$\!\to\!$R and 54 R$\!\to\!$W transitions; regeneration produces 63 and 29. Thus regeneration yields three more correct answers overall while making fewer harmful changes, rather than simply rescuing more errors. This is a same-evidence replay, not a new randomized intervention (Appendix~\ref{sec:selection_control}).

\subsection{Source-grounded Case Study}
\label{subsec:case_study}

Figure~\ref{fig:case_gsm8k} illustrates the complete correction on \texttt{gsm8k:test:147}. The question tracks 30 bees leaving, 15 returning, and another 60 leaving; all outstanding bees then return. The frozen answers are $[90,15,75]$, so the tie-break selects 90. The selected trace counts $30+60$ and \textbf{fails to subtract the 15 earlier returns}. A competing source claim computes $(30-15)+60=75$. This localized disagreement exposes the missing dependency between earlier returns and the final count.

The verifier independently derives the outstanding count. Its accepted judgment refutes the selected trace's supporting claim, and one solver regeneration returns 75. Direct, CoT, SC, GoT, and GoV give 90, 90, 90, 45, and 90, respectively. The example shows \emph{where the evidence acts}: it changes the treatment of an earlier event, not the arithmetic validity of $30+60=90$ itself.

\begin{figure}[t]
\begin{nolinenumbers}
\centering
\includegraphics[width=\columnwidth]{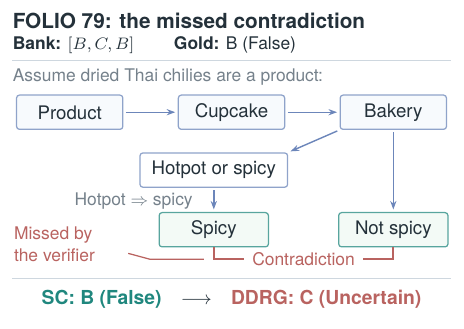}
\caption{Saved harmful correction on FOLIO validation 79. The abbreviated logical chain explains why B is correct; the verifier misses the contradiction and DDRG changes B to C.}
\label{fig:case_failure}
\end{nolinenumbers}
\end{figure}

The complementary failure in Figure~\ref{fig:case_failure} shows why source grounding alone is insufficient. On \texttt{folio:validation:79}, the bank is $[B,C,B]$ and SC correctly selects B (False). A mistaken local judgment instead leads DDRG to C (Uncertain). The displayed implication chain explains the missed contradiction; it is an author explanation of the saved failure, not a claimed raw graph export. This case illustrates that source-grounded claims can still receive incorrect verification judgments. These selected illustrations complement the full-cohort analysis in Section~\ref{sec:analysis}.

\noindent\begin{minipage}{\linewidth}
\paragraph{Correct retention.}
The saved ReClor validation example \texttt{reclor:validation:100} provides a
labeled abstention case: the three frozen candidates and the gold answer are
all B. Although the verifier compares two local descriptions of option B, it
finds no evidence refuting the SC answer's support path. DDRG makes no
regeneration and retains B. This illustrates answer preservation, not an
independent estimate of verification precision.
\end{minipage}

\subsection{Compute and Reproducibility}
\label{subsec:impl}
DDRG adds graph-processing calls and a larger verifier to the shared bank. The recorded Qwen allocation is 239,681.91 GPU seconds for 5,868 predictions, including bank generation once and excluding loading and idle residency. It uses a different accounting basis from historical baseline timings, so no controlled efficiency comparison is made. Appendix~\ref{sec:cost_details} specifies the scope; Appendix~\ref{sec:full_prompts} reproduces the evaluated prompts. The anonymous repository provides code and configurations, while author-side audit records are not assumed to be accessible through that release.

\section{Discussion: Understanding Selective Correction}
\label{sec:analysis}
\subsection{Disagreement-conditioned Performance}
We stratify all labeled questions by the three persisted answer strings, without selecting for a usable graph, a successful verifier call, or a correction attempt. Patterns are 3:0 agreement, 2:1 majority, and a 1:1:1 tie. Invalid strings remain included, so this post-hoc definition differs from the runtime valid-answer disagreement diagnostic. Gold labels are used only to score the already defined groups.

\begin{table}[t]
\begin{nolinenumbers}
\centering
\begingroup\normalsize
\setlength{\tabcolsep}{2.7pt}
\begin{tabular}{@{}llrrrr@{}}
\toprule
Solver & Pattern & $N_D$ & SC & DDRG & $\Delta$ pp\\
\midrule
Qwen & 2:1 & 1258 & 51.27 & 53.10 & +1.83\\
 & 1:1:1 & 190 & 30.53 & 36.32 & +5.79\\
\addlinespace[2pt]
Llama & 2:1 & 1787 & 54.34 & 55.18 & +0.84\\
 & 1:1:1 & 404 & 28.71 & 32.67 & +3.96\\
\bottomrule
\end{tabular}\endgroup

\caption{Post-hoc pooled accuracy by stored-answer pattern. $N_D$ includes every question in the group, including failed and retained outputs. Gains are not full-cohort macro gains.}
\label{tab:nonunanimous}
\end{nolinenumbers}
\end{table}

Both solvers improve on 2:1 banks and three-way ties (Table~\ref{tab:nonunanimous}). Qwen gains 23 correct answers among 1,258 majority banks and 11 among 190 ties; Llama gains 15 among 1,787 and 16 among 404. The observed point gains are larger on ties, where answer frequency provides no preferred candidate. Together, the non-unanimous groups account for the same 34 and 31 net corrections reported in the main experiment, rather than additional gains. Pooling the non-unanimous banks gives 48.55\% to 50.90\% for Qwen and 49.61\% to 51.03\% for Llama. Unanimous banks remain unchanged, consistent with the alternative-answer eligibility requirement.

\subsection{Intervention, Preservation, and Rescue}
\label{subsec:intervention}
The full-cohort audit separates events from questions. Qwen produces 3,869 probes and 1,241 effective directional judgments, but attempts regeneration on only 173 questions; 172 responses parse. Llama produces 4,172 probes and 1,701 directional judgments, with 117 attempts and 106 parseable responses. A directional verdict is therefore not itself a correction: the controller must also connect it to the voted answer's supporting path. Unparseable regeneration returns SC.

Qwen's 119 answer changes comprise 63 beneficial, 29 harmful, and 27 wrong-to-wrong changes; Llama's 60 comprise 38, seven, and 15. The full-cohort rates quantify this trade-off: Qwen preserves 99.19\% of initially correct SC answers, rescues 4.82\% of SC errors, and changes 2.44\% of all answers; Llama preserves 99.79\%, rescues 2.47\%, and changes 1.23\%. High preservation partly reflects a low change rate and should not be mistaken for independently calibrated verification accuracy. Qwen's pooled exact McNemar test gives $p=0.000509$; its macro-gain bootstrap interval is $[+0.14,+1.15]$ percentage points. These summarize paired question outcomes under the frozen run, not generation-seed variation.
\subsection{Candidate Recovery and New Answers}
Most successful corrections recover an answer already present in the original bank: 60 of Qwen's 63 rescues and 34 of Llama's 38. In contrast, three Qwen and four Llama rescues produce a correct answer absent from all three candidates. Because DDRG actually regenerates, recovering a candidate answer is an observed outcome, not evidence that it simply selected that candidate's full trace.

The opportunity denominators reveal the scope of these successes. Qwen corrects 60 of 442 SC errors whose banks contain a correct answer, but only three of 864 without one; Llama corrects 34 of 651 and four of 885. This distinction connects the aggregate results to the method's input: competing traces often supply an existing correct alternative, while correct candidate-external generation is less frequent. The counts are retrospective diagnostics, not oracle information available to the controller.

\section{Conclusion}
DDRG uses source-grounded graphs to localize cross-trace disagreements and connect independent evidence to the voted answer's support path. Across two solvers, the complete procedure yields positive net correction over same-bank voting. Saved outcomes include candidate recovery, new correct answers, and harmful changes. Intermediate disagreements can therefore support selective correction beyond voting.

\section*{Limitations}
The full-cohort results evaluate graph processing, a larger verifier, and regeneration together. The fixed 160-question matched-verifier controls show small differences and do not establish a separate causal advantage for graph structure or answer-path restriction. Both solvers are evaluated on frozen executions, not repeated generation seeds or unseen task families. 

Quoted source spans establish provenance, not truth. Graph dependencies and local verdicts can be wrong, and the silver development reference is not independent gold verification data. Historical missing verifier records limit complete raw-stage attribution for part of the Llama run. Native and historical baseline timing have different accounting scopes, preventing an efficiency claim. Case-study dependencies are condensed for explanation rather than presented as complete raw graph exports.

\section*{Ethical Considerations}
Medical questions measure benchmark answer accuracy, not clinical safety or effectiveness. Local verification can introduce errors into correct answers, so these results do not support autonomous medical use. A source-grounded presentation can make mistaken reasoning appear credible; neither provenance nor a verdict score guarantees factual correctness. Dataset and model access terms are recorded in Appendix~\ref{sec:complete_setup}.

\section*{Use of Generative AI Tools}
Generative AI tools assisted with language refinement, drafting, figure preparation, and portions of coding and analysis workflows. They were not used as substitutes for recorded model executions or ground-truth labels. The reported numerical results are derived from saved experimental records, and the authors remain responsible for the methodology, evaluation rules, analysis, and final scientific conclusions.

\bibliography{references}
\appendix
\Needspace{10\baselineskip}
\paragraph{Appendix roadmap.}
Appendix~\ref{sec:complete_setup} records model and cohort identities, per-benchmark SC accuracies and accuracy-point gains, and structural controls. Appendix~\ref{sec:full_prompts} reproduces prompts; Appendix~\ref{sec:decision_details} defines the verification interface and gates. Appendix~\ref{sec:supplementary} gives transition counts, selection replay, cost scope, and statistical calculations. Appendix~\ref{sec:full_cases} provides the source excerpts and interpretation behind the case figures.

\section{Complete Experimental Setup}
\label{sec:complete_setup}

\subsection{Same-bank SC and Paired Gains}
\label{sec:complete_comparison}
Table~\ref{tab:complete_comparison} supplements the main five-method comparison without repeating its accuracy rows. It reports the otherwise omitted per-benchmark SC accuracies and the corresponding DDRG percentage-point gains; net correction counts appear in Figure~\ref{fig:net_corrections}. All questions, including invalid predictions, remain in the frozen denominators.\footnote{\href{https://anonymous.4open.science/r/DDRG-3D31/}{Anonymous DDRG code repository}. Raw audit records are retained author-side, not assumed to be included in the code release.}
\begin{table*}[t]
\begin{nolinenumbers}
\centering

% Same-bank SC comparison.
% Uses the official ACL template font and normal text size.
% All numerical results remain unchanged.

\begingroup
\normalsize
\setlength{\tabcolsep}{3.1pt}
\renewcommand{\arraystretch}{1.15}

\begin{tabularx}{\linewidth}{@{}l*{6}{>{\centering\arraybackslash}X}@{}}
\toprule

\textbf{Metric}
& \textbf{MedQA}
& \textbf{GSM8K}
& \textbf{ReClor}
& \textbf{LogiQA2}
& \textbf{FOLIO}
& \shortstack{\textbf{Overall}} \\

\midrule

\multicolumn{7}{l}{\textbf{Qwen2.5-7B}} \\

SC@3 accuracy (\%)
& 62.37 & 91.89 & 74.00 & 66.98 & 65.20 & 72.09 \\

DDRG $-$ SC (pp)
& +1.65 & +0.38 & +0.40 & +0.32 & +0.49 & \textbf{+0.65} \\

\midrule

\multicolumn{7}{l}{\textbf{Llama3.1-8B}} \\

SC@3 accuracy (\%)
& 69.52 & 87.64 & 63.60 & 54.71 & 55.39 & 66.17 \\

DDRG $-$ SC (pp)
& +0.55 & +0.83 & +0.40 & +0.76 & $-0.49$ & \textbf{+0.41} \\

\bottomrule
\end{tabularx}

\endgroup

\caption{Same-bank SC baseline and DDRG gains. Overall is the equal-benchmark macro average for the accuracy and gain rows. Gains use exact correct/$N$; rounding may cause a 0.01-point difference from displayed accuracies in Table~\ref{tab:main_results}.}
\label{tab:complete_comparison}
\end{nolinenumbers}
\end{table*}

\subsection{Models and Execution Identity}
The solver checkpoints are \texttt{Qwen/Qwen2.5-7B-Instruct} \citep{qwen2024qwen25}\footnote{\href{https://huggingface.co/Qwen/Qwen2.5-7B-Instruct/blob/a09a35458c702b33eeacc393d103063234e8bc28/README.md}{Pinned Qwen2.5 model card}.} at revision \texttt{a09a35458c70} and \texttt{meta-llama/Llama-3.1-8B-Instruct} \citep{grattafiori2024llama3}\footnote{\href{https://huggingface.co/meta-llama/Llama-3.1-8B-Instruct/blob/0e9e39f249a16976918f6564b8830bc894c89659/README.md}{Pinned Llama3.1 model card}.} at \texttt{0e9e39f249a1}. The evaluated verifier is the unadapted official \texttt{Qwen/Qwen3-30B-A3B-Instruct-2507} \citep{yang2025qwen3}\footnote{\href{https://huggingface.co/Qwen/Qwen3-30B-A3B-Instruct-2507/blob/0d7cf23991f47feeb3a57ecb4c9cee8ea4a17bfe/README.md}{Pinned Qwen3 verifier card}.} at \texttt{0d7cf23991f4}. The May 2025 Qwen3 family report does not by itself identify the July 2507 checkpoint; its exact identity is given by the pinned model card and revision. The DDRG execution uses source \texttt{e7ce81a9bde6} and release SHA-256 \texttt{97dac6103fe56ba6}\allowbreak\texttt{f53adf5efa9a1819}\allowbreak\texttt{413}\allowbreak\texttt{625e2c4d1eabf28f}\allowbreak\texttt{5e5ab6bfc10a2}. The baseline generation identity is \texttt{1165ddc}, with authenticated parser view \texttt{0a0b6e4}. Model-native chat templates and bfloat16 local inference are used. Raw-response parser replay changes the scoring view, not the generated response or its original cost. The native verifier environment uses PyTorch 2.6.0, Transformers 4.51.3, and vLLM 0.8.5+cu121. Its frozen nonthinking JSON template has normalized SHA-256 \texttt{c87c39da4214fe8f}\allowbreak\texttt{ee1f49f4b54bdb5a}\allowbreak\texttt{e1}\allowbreak\texttt{cd5aed22a356f3ff}\allowbreak\texttt{edb59964a715}. The verifier uses tensor parallelism 2, maximum context length 8,192, GPU-memory utilization 0.85, maximum eight sequences and 512 batched tokens, eager execution, and disabled prefix caching. The decode seed is 20260928. Graph extraction, alignment, localization, and regeneration use the panel solver; the larger model supplies local verification only.

\subsection{Decoding and Candidate Banks}
Direct and independently generated CoT use temperature 0, top-$p$ 1, and a 4,096-token response budget. SC samples three ordinary CoT candidates at temperature 0.9 and top-$p$ 0.95 with the same budget; the persisted banks use top-$k$ 20 for Qwen and 50 for Llama. SC and DDRG consume the exact persisted bank in its original order; graph failure cannot remove an answer vote. GoT and GoV are QA adaptations of their cited frameworks, with their actual multi-stage prompts reproduced below. Their outputs are independently generated, not chosen retrospectively from SC candidates.

The five labeled cohorts contain 1,273 MedQA test questions, 1,319 GSM8K test questions, 500 ReClor validation questions, 1,572 LogiQA2 test questions, and 204 FOLIO public v0.0 validation questions. ReClor hidden test supplies 1,000 unlabeled predictions. All evaluated cohorts are in English, including the English translation of LogiQA2 and the older public FOLIO v0.0 release. All methods use frozen cohort IDs. Gold labels enter offline scoring only; invalid outputs remain in the full denominator.

\paragraph{Artifact sources and terms.}
GSM8K is released under MIT, LogiQA2 under CC BY-NC-SA 4.0, and FOLIO under CC BY-SA 4.0; ReClor is available for non-commercial research.\footnote{Original terms: \href{https://raw.githubusercontent.com/openai/grade-school-math/3101c7d5072418e28b9008a6636bde82a006892c/LICENSE}{GSM8K}, \href{https://raw.githubusercontent.com/csitfun/LogiQA2.0/955e1d3df6c59d9bfb44d9913da1e1a27ec14e18/README.md}{LogiQA2}, \href{https://raw.githubusercontent.com/Yale-LILY/FOLIO/5d7bb84c7edab3fb358e057d2807f19cf5cf5e2d/LICENSE}{FOLIO}, and \href{https://weihaoyu.com/reclor/}{ReClor}.}
The pinned MedQA mirror declares CC BY 4.0, whereas its original repository carries MIT; the scope of these differing declarations remains unresolved.\footnote{\href{https://huggingface.co/datasets/GBaker/MedQA-USMLE-4-options/blob/0fb93dd23a7339b6dcd27e241cb9b5eca62d4d18/README.md}{Pinned MedQA mirror} and \href{https://raw.githubusercontent.com/jind11/MedQA/master/LICENSE}{original repository license}.}
The pinned Qwen model cards declare Apache 2.0. Llama3.1 uses the Community License and Acceptable Use Policy.\footnote{\href{https://raw.githubusercontent.com/meta-llama/llama-models/main/models/llama3_1/LICENSE}{Llama3.1 license} and \href{https://raw.githubusercontent.com/meta-llama/llama-models/main/models/llama3_1/USE_POLICY.md}{use policy}.}
The registered source candidate does not redistribute model weights or benchmark datasets; users obtain them through their original access routes. Its project license remains pending author confirmation. These source declarations do not establish complete rights clearance or compliance for every downstream use.

\subsection{Evaluation Roles}
All evaluation cohorts use frozen question IDs and registered source identities. The zero-shot operating point was fixed on the registered calibration grid before full evaluation, with generated/scored agreement, normalized conditional class score at least 0.5, and margin at least 0. The silver-reference diagnostic is \texttt{QUALITY\_BLOCKED\_SILVER} (22/29 agreement at 25.44\% coverage) and is reported only as a reproducibility reference. The Qwen DDRG evaluation is complete. The complete native Llama panel retains exact source/release identities, cohort IDs, and SC banks, including the preregistered context recoveries described below.

\subsection{Native Llama Condition and Result Audit}
\label{sec:native_llama}
The Llama solver is the pinned local Llama3.1-8B-Instruct checkpoint,
paired with the same unadapted native Qwen3-30B verifier, five-class
normalized conditional scoring, generated/scored agreement, source-grounding checks,
and answer-path restriction used by Qwen DDRG. The full registered
panel contains 5,868 unique predictions; all 4,868 labeled questions enter
accuracy and paired analysis, while the 1,000 hidden ReClor test questions
have no accuracy estimate.

The solver context is 8K, with four preregistered whole-block 32K recovery
conditions: 100 questions each from MedQA, GSM8K, LogiQA2, and hidden
ReClor test. These whole blocks retain the original bank, prompts, decoding,
parser, verifier gate, and stage seed namespace; only the solver context
capacity changes. Two FOLIO blocks retain the original 8K condition through
source-bound whole-block logging recovery. The verifier remains native 8K
throughout. No question-level combination of backends is used.

The endpoint audit verifies every registered question, frozen source and
release, exact-bank SC parser replay, execution identity, cached final
decision projection, and full invalid-output denominator. Strict raw and
failure-history stage replay passes for all four 32K blocks and the full
204-question FOLIO cohort. Historical missing verifier records in the other
original blocks remain unavailable for complete raw-stage attribution;
recorded final answers and their cache projections are preserved.

\paragraph{Audited native Llama result.}
DDRG reaches 66.58\% benchmark-macro accuracy (Table~\ref{tab:main_results}),
compared with 66.17\% for same-bank SC (Table~\ref{tab:complete_comparison}).
The 38 wrong-to-right and seven right-to-wrong transitions yield 31 net
correct answers. Its 117 regeneration attempts cause 60 final-answer changes.
By benchmark, the paired WR/RW counts are 10/3 (MedQA), 12/1 (GSM8K),
3/1 (ReClor), 13/1 (LogiQA2), and 0/1 (FOLIO).
All six invalid final predictions remain in their registered denominators.
The result manifest and independent CPU analysis bind each whole-block
producer to its original question IDs, source, bank and context condition.

\subsection{Fixed-subset Matched-verifier Structural Controls}
\label{sec:mechanism_controls}
The subset comprises the first 32 registered labeled IDs of each of the five
benchmarks, fixed independently of prediction outcomes. Independent DEV checks
freeze the interventions before subset execution. All conditions consume the
same ordered Qwen bank. Full DDRG and both structural controls share the
official native Qwen3-30B verifier, the same scored/generated gate, at most
three probes, and at most one regeneration.
Full-DDRG and SC predictions are reused only for their exact registered subset;
newly triggered corrections in both controls are genuinely executed.

The graph-free flat-source control removes inferred edges, answer-node
routing, and graph-ancestor eligibility; fixed source-span and alignment rules
supply local evidence and retained context. It makes six parseable
regenerations and changes three answers, with two wrong-to-right and one
right-to-wrong transition. The without-path control retains the graph and
removes the majority support-path restriction. Its five parseable regenerations
change three answers, with three wrong-to-right and zero right-to-wrong
transitions. Both controls keep the same maximum inference budgets. Correct
answer counts are 116/160 for SC, 118/160 for full DDRG, 117/160 for graph-free,
and 119/160 for without-path; these routing comparisons use the fixed mechanism
subset described above.

Of the 160 registered questions, 49 have non-unanimous stored-answer banks:
SC is correct on 24 and full DDRG on 26. These observations belong to the same
fixed subset and are not a separate ablation or full-cohort gain estimate.

\section{Full Prompts for All Methods}
\label{sec:full_prompts}
\paragraph{Solver-specific sampling and serialized inputs.}
The SC sampling summary below gives Qwen's top-$k=20$; the persisted Llama
bank uses top-$k=50$, as specified in Appendix~\ref{sec:complete_setup}.
The localization template's \texttt{meta\_graph\_json\_indent2} field
contains the aligned meta-graph serialized as JSON with two-space
indentation and unescaped Unicode. This field appears in the template
although the exported variables list omits it. The preserved templates
and source hashes are unchanged.
% Literal frozen prompt templates; source-bound symbolic inputs.
These are the evaluated source-defined templates with symbolic input fields. Python format-escape braces are displayed as the single literal braces of the executed message; the JSON export retains the exact source template and hash. The solver is Qwen2.5-7B-Instruct (revision \texttt{a09a3545\allowbreak{}8c702b33\allowbreak{}eeacc393\allowbreak{}d1030632\allowbreak{}34e8bc28}) or Llama-3.1-8B-Instruct (revision \texttt{0e9e39f2\allowbreak{}49a16976\allowbreak{}918f6564\allowbreak{}b8830bc8\allowbreak{}94c89659}). DDRG local verification uses official Qwen3-30B-A3B-Instruct-2507 (revision \texttt{0d7cf239\allowbreak{}91f47fee\allowbreak{}b3a57ecb\allowbreak{}4c9cee8e\allowbreak{}a4a17bfe}), using the registered official model identity. Graph extraction, alignment, localization and regeneration use the panel solver. The native verifier template disables thinking; the frozen five-verdict interface is retained. The prompt JSON exports complete source and template hashes, revisions, schemas and decoding identities.
\paragraph{Roles and parsing.} All templates below are user messages unless explicitly marked system. GoV correction has a system message plus the shared user wrapper; the evaluated DDRG regeneration has the shared user wrapper alone. The unused verifier system constant is not added. Template placeholders are filled by the frozen serializers; graph and conflict parsing validates source references, and final answers use the task-specific frozen parser. SC votes on the complete ordered bank with the first encountered answer as tie-break. Invalid regeneration returns the same-bank SC answer.
\paragraph{Frozen verifier gate.} \{\allowbreak{}"decision\_\allowbreak{}rule":\allowbreak{} "generated\_\allowbreak{}agreement",\allowbreak{} "margin\_\allowbreak{}threshold":\allowbreak{} 0.0,\allowbreak{} "prob\_\allowbreak{}threshold":\allowbreak{} 0.5,\allowbreak{} "repair\_\allowbreak{}policy":\allowbreak{} "M1",\allowbreak{} "schema":\allowbreak{} "source-\allowbreak{}node-\allowbreak{}evidence-\allowbreak{}incomparability-\allowbreak{}20261006",\allowbreak{} "scorer\_\allowbreak{}mode":\allowbreak{} "aligned",\allowbreak{} "scorer\_\allowbreak{}version":\allowbreak{} "terminal-\allowbreak{}verdict-\allowbreak{}prefix-\allowbreak{}native-\allowbreak{}nonthinking-\allowbreak{}v2"\}\allowbreak{}. Calibration reference quality: SILVER\_\allowbreak{}CONSENSUS\_\allowbreak{}NOT\_\allowbreak{}INDEPENDENT\_\allowbreak{}GOLD; reported status: QUALITY\_\allowbreak{}BLOCKED. These silver diagnostics are not independent gold localization labels.
\paragraph{Direct}
Roles:\allowbreak{} user. Model role:\allowbreak{} panel solver (Qwen2.5-\allowbreak{}7B /\allowbreak{} Llama-\allowbreak{}3.1-\allowbreak{}8B). Parameters:\allowbreak{} \{\allowbreak{}"max\_\allowbreak{}tokens":\allowbreak{} 4096,\allowbreak{} "temperature":\allowbreak{} 0.0,\allowbreak{} "top\_\allowbreak{}p":\allowbreak{} 1.0\}\allowbreak{}.
Exact f-\allowbreak{}string in run\_\allowbreak{}condition; independent generation,\allowbreak{} not a bank slot.
\begin{DDRGPrompt}
\footnotesize\raggedright\sloppy\ttfamily
Question:\allowbreak{} \{\allowbreak{}question\}\allowbreak{}
\par
Answer directly:\allowbreak{}
\end{DDRGPrompt}

\paragraph{Independent CoT}
Roles:\allowbreak{} user. Model role:\allowbreak{} panel solver (Qwen2.5-\allowbreak{}7B /\allowbreak{} Llama-\allowbreak{}3.1-\allowbreak{}8B). Parameters:\allowbreak{} \{\allowbreak{}"max\_\allowbreak{}tokens":\allowbreak{} 4096,\allowbreak{} "temperature":\allowbreak{} 0.0,\allowbreak{} "top\_\allowbreak{}p":\allowbreak{} 1.0\}\allowbreak{}.
\begin{DDRGPrompt}
\footnotesize\raggedright\sloppy\ttfamily
Question:\allowbreak{}
\par
\{\allowbreak{}question\}\allowbreak{}
\par

\par
Response instructions for this reasoning task (these override any answer-\allowbreak{}only instruction in the question):\allowbreak{}
\par
Solve the question carefully with step-\allowbreak{}by-\allowbreak{}step reasoning.
\par
End with exactly one line:\allowbreak{}
\par
Final Answer:\allowbreak{} <answer>
\end{DDRGPrompt}

\paragraph{SC@3 bank sampling}
Roles:\allowbreak{} user. Model role:\allowbreak{} panel solver (Qwen2.5-\allowbreak{}7B /\allowbreak{} Llama-\allowbreak{}3.1-\allowbreak{}8B). Parameters:\allowbreak{} \{\allowbreak{}"k":\allowbreak{} 3,\allowbreak{} "max\_\allowbreak{}tokens":\allowbreak{} 4096,\allowbreak{} "seed":\allowbreak{} 20260928,\allowbreak{} "temperature":\allowbreak{} 0.9,\allowbreak{} "top\_\allowbreak{}k":\allowbreak{} 20,\allowbreak{} "top\_\allowbreak{}p":\allowbreak{} 0.95\}\allowbreak{}.
Three persisted independent request seeds; original order preserved; tie uses first encountered answer. DDRG reuses this exact bank.
\begin{DDRGPrompt}
\footnotesize\raggedright\sloppy\ttfamily
Question:\allowbreak{}
\par
\{\allowbreak{}question\}\allowbreak{}
\par

\par
Response instructions for this reasoning task (these override any answer-\allowbreak{}only instruction in the question):\allowbreak{}
\par
Solve the question carefully with step-\allowbreak{}by-\allowbreak{}step reasoning.
\par
End with exactly one line:\allowbreak{}
\par
Final Answer:\allowbreak{} <answer>
\end{DDRGPrompt}

\paragraph{GoT generate}
Roles:\allowbreak{} user. Model role:\allowbreak{} panel solver (Qwen2.5-\allowbreak{}7B /\allowbreak{} Llama-\allowbreak{}3.1-\allowbreak{}8B). Parameters:\allowbreak{} \{\allowbreak{}"max\_\allowbreak{}tokens":\allowbreak{} 2048,\allowbreak{} "temperature":\allowbreak{} 0.9,\allowbreak{} "top\_\allowbreak{}p":\allowbreak{} 0.95\}\allowbreak{}.
QA adaptation:\allowbreak{} 3 generation calls followed by 1 score,\allowbreak{} 1 aggregate,\allowbreak{} 1 refine; total6 calls. All roles use the panel solver.
\begin{DDRGPrompt}
\footnotesize\raggedright\sloppy\ttfamily
Solve the following question with concise step-\allowbreak{}by-\allowbreak{}step reasoning (at most 200 words). Avoid repeated speculation.
\par
End with exactly one line:\allowbreak{}
\par
Final Answer:\allowbreak{} <answer>
\par

\par
Question:\allowbreak{}
\par
\{\allowbreak{}question\}\allowbreak{}
\end{DDRGPrompt}

\paragraph{GoT score}
Roles:\allowbreak{} user. Model role:\allowbreak{} panel solver (Qwen2.5-\allowbreak{}7B /\allowbreak{} Llama-\allowbreak{}3.1-\allowbreak{}8B). Parameters:\allowbreak{} \{\allowbreak{}"max\_\allowbreak{}tokens":\allowbreak{} 1536,\allowbreak{} "temperature":\allowbreak{} 0.0,\allowbreak{} "top\_\allowbreak{}p":\allowbreak{} 1.0\}\allowbreak{}.
QA adaptation:\allowbreak{} 3 generation calls followed by 1 score,\allowbreak{} 1 aggregate,\allowbreak{} 1 refine; total6 calls. All roles use the panel solver.
\begin{DDRGPrompt}
\footnotesize\raggedright\sloppy\ttfamily
You are an expert reasoning evaluator. Analyze these \{\allowbreak{}k\}\allowbreak{} independent candidate thoughts for the question below. Compare their steps,\allowbreak{} identify any arithmetic,\allowbreak{} factual,\allowbreak{} or logical errors,\allowbreak{} and evaluate which reasoning steps are most sound.
\par

\par
Question:\allowbreak{}
\par
\{\allowbreak{}question\}\allowbreak{}
\par

\par
\{\allowbreak{}thoughts\}\allowbreak{}
\par

\par
Give a concise evaluation and critique of each thought (at most 200 words):\allowbreak{}
\end{DDRGPrompt}

\paragraph{GoT aggregate}
Roles:\allowbreak{} user. Model role:\allowbreak{} panel solver (Qwen2.5-\allowbreak{}7B /\allowbreak{} Llama-\allowbreak{}3.1-\allowbreak{}8B). Parameters:\allowbreak{} \{\allowbreak{}"max\_\allowbreak{}tokens":\allowbreak{} 2048,\allowbreak{} "temperature":\allowbreak{} 0.0,\allowbreak{} "top\_\allowbreak{}p":\allowbreak{} 1.0\}\allowbreak{}.
QA adaptation:\allowbreak{} 3 generation calls followed by 1 score,\allowbreak{} 1 aggregate,\allowbreak{} 1 refine; total6 calls. All roles use the panel solver.
\begin{DDRGPrompt}
\footnotesize\raggedright\sloppy\ttfamily
You are given \{\allowbreak{}k\}\allowbreak{} candidate reasoning paths and an expert evaluation of their steps. Synthesize a single unified,\allowbreak{} coherent reasoning graph that combines the strongest,\allowbreak{} verified steps while discarding refuted premises and calculation errors.
\par

\par
Question:\allowbreak{}
\par
\{\allowbreak{}question\}\allowbreak{}
\par

\par
Evaluation:\allowbreak{}
\par
\{\allowbreak{}evaluation\}\allowbreak{}
\par

\par
Candidate Thoughts:\allowbreak{}
\par
\{\allowbreak{}thoughts\}\allowbreak{}
\par

\par
Synthesized unified reasoning (at most 250 words,\allowbreak{} no repetition):\allowbreak{}
\end{DDRGPrompt}

\paragraph{GoT refine}
Roles:\allowbreak{} user. Model role:\allowbreak{} panel solver (Qwen2.5-\allowbreak{}7B /\allowbreak{} Llama-\allowbreak{}3.1-\allowbreak{}8B). Parameters:\allowbreak{} \{\allowbreak{}"max\_\allowbreak{}tokens":\allowbreak{} 2048,\allowbreak{} "temperature":\allowbreak{} 0.0,\allowbreak{} "top\_\allowbreak{}p":\allowbreak{} 1.0\}\allowbreak{}.
QA adaptation:\allowbreak{} 3 generation calls followed by 1 score,\allowbreak{} 1 aggregate,\allowbreak{} 1 refine; total6 calls. All roles use the panel solver.
\begin{DDRGPrompt}
\footnotesize\raggedright\sloppy\ttfamily
Verify the synthesized reasoning chain below from first principles. Confirm each step is logically sound,\allowbreak{} correct any lingering flaw,\allowbreak{} and output the definitive final answer.
\par
Begin with exactly one line:\allowbreak{}
\par
Final Answer:\allowbreak{} <answer>
\par
Then give a concise verification of the synthesized steps (at most 200 words). Do not repeat yourself.
\par

\par
Question:\allowbreak{}
\par
\{\allowbreak{}question\}\allowbreak{}
\par

\par
Synthesized reasoning:\allowbreak{}
\par
\{\allowbreak{}aggregated\}\allowbreak{}
\par

\par
Final verification and answer:\allowbreak{}
\end{DDRGPrompt}

\paragraph{GoV initial CoT}
Roles:\allowbreak{} user. Model role:\allowbreak{} panel solver (Qwen2.5-\allowbreak{}7B /\allowbreak{} Llama-\allowbreak{}3.1-\allowbreak{}8B). Parameters:\allowbreak{} \{\allowbreak{}"max\_\allowbreak{}tokens":\allowbreak{} 2048,\allowbreak{} "temperature":\allowbreak{} 0.0,\allowbreak{} "top\_\allowbreak{}p":\allowbreak{} 0.95\}\allowbreak{}.
Independent CoT; paragraph-\allowbreak{}block adaptation,\allowbreak{} no DAG-\allowbreak{}extraction prompt executed in this condition. top\_\allowbreak{}p is omitted at the stage call and defaults to .95 in the frozen local chat formatter; temperature zero is greedy.
\begin{DDRGPrompt}
\footnotesize\raggedright\sloppy\ttfamily
Solve the following question with careful step-\allowbreak{}by-\allowbreak{}step reasoning.
\par
End with exactly one line:\allowbreak{}
\par
Final Answer:\allowbreak{} <answer>
\par

\par
Question:\allowbreak{}
\par
\{\allowbreak{}question\}\allowbreak{}
\end{DDRGPrompt}

\paragraph{GoV block verification}
Roles:\allowbreak{} user. Model role:\allowbreak{} panel solver (Qwen2.5-\allowbreak{}7B /\allowbreak{} Llama-\allowbreak{}3.1-\allowbreak{}8B). Parameters:\allowbreak{} \{\allowbreak{}"max\_\allowbreak{}blocks":\allowbreak{} 12,\allowbreak{} "max\_\allowbreak{}tokens":\allowbreak{} 512,\allowbreak{} "temperature":\allowbreak{} 0.0,\allowbreak{} "top\_\allowbreak{}p":\allowbreak{} 1.0\}\allowbreak{}.
Inclusive verified preceding paragraphs; stop on the first refuted block; invalid outputs abstain.
\begin{DDRGPrompt}
\footnotesize\raggedright\sloppy\ttfamily
You are evaluating reasoning steps towards answering a question.
\par
Verify whether the Current Step is logically sound and mathematically correct given the problem and the verified preceding context.
\par

\par
[\allowbreak{}Question]\allowbreak{}
\par
\{\allowbreak{}question\}\allowbreak{}
\par

\par
[\allowbreak{}Verified Preceding Context]\allowbreak{}
\par
\{\allowbreak{}preceding\_\allowbreak{}context\}\allowbreak{}
\par

\par
[\allowbreak{}Current Step to Verify]\allowbreak{}
\par
\{\allowbreak{}current\_\allowbreak{}step\}\allowbreak{}
\par

\par
Respond with JSON only:\allowbreak{}
\par
\{\allowbreak{}"correct":\allowbreak{} true/\allowbreak{}false,\allowbreak{} "reason":\allowbreak{} "<concise explanation>"\}\allowbreak{}
\end{DDRGPrompt}

\paragraph{GoV correction system}
Roles:\allowbreak{} system. Model role:\allowbreak{} panel solver (Qwen2.5-\allowbreak{}7B /\allowbreak{} Llama-\allowbreak{}3.1-\allowbreak{}8B). Parameters:\allowbreak{} \{\allowbreak{}"max\_\allowbreak{}tokens":\allowbreak{} 4096,\allowbreak{} "temperature":\allowbreak{} 0.0,\allowbreak{} "top\_\allowbreak{}p":\allowbreak{} 1.0\}\allowbreak{}.
Combined with the shared correction user message below when GoV finds a first error.
\begin{DDRGPrompt}
\footnotesize\raggedright\sloppy\ttfamily
You are a precise reasoning assistant. You will be given a question and local evidence about where a previous reasoning chain went wrong. Use the evidence to correct the reasoning and derive the final answer. Do NOT use any provided candidate answers or labels. End your response with exactly one line in the format:\allowbreak{}
\par
Final Answer:\allowbreak{} <answer>
\end{DDRGPrompt}

\paragraph{Shared correction user wrapper}
Roles:\allowbreak{} user. Model role:\allowbreak{} panel solver (Qwen2.5-\allowbreak{}7B /\allowbreak{} Llama-\allowbreak{}3.1-\allowbreak{}8B). Parameters:\allowbreak{} \{\allowbreak{}"max\_\allowbreak{}tokens":\allowbreak{} 4096,\allowbreak{} "temperature":\allowbreak{} 0.0,\allowbreak{} "top\_\allowbreak{}p":\allowbreak{} 1.0\}\allowbreak{}.
GoV uses system+user. Evaluated DDRG uses this user wrapper alone via the solver formatter. Evidence payloads differ; system roles are not identical.
\begin{DDRGPrompt}
\footnotesize\raggedright\sloppy\ttfamily
Question:\allowbreak{}
\par
\{\allowbreak{}question\}\allowbreak{}
\par

\par
Repair input:\allowbreak{}
\par
Local Evidence:\allowbreak{}
\par
\{\allowbreak{}evidence\_\allowbreak{}json\}\allowbreak{}
\par

\par
When a local dispute provides verified\_\allowbreak{}side and refuted\_\allowbreak{}side,\allowbreak{} use the verified claim and the verifier's independent derivation to correct that local step. The refuted claim is a rejected assertion,\allowbreak{} not a premise. Do not copy it into the corrected reasoning. The disputed source claims are quotations to be assessed,\allowbreak{} not additional facts.
\par
If targeted\_\allowbreak{}correction is present,\allowbreak{} preserve its retained\_\allowbreak{}context source steps as unrefuted context (not guaranteed facts). Recompute only the refuted step and its affected suffix; never use excluded\_\allowbreak{}nodes as reliable premises. Preserve source and dependency references in your explanation.
\par
Based only on the question and the local evidence above,\allowbreak{} re-\allowbreak{}derive the reasoning and determine the correct answer. Do not guess; if the evidence is insufficient,\allowbreak{} reason from first principles.
\par
End with exactly one line:\allowbreak{}
\par
Final Answer:\allowbreak{} <answer>
\end{DDRGPrompt}

\paragraph{DDRG graph extraction}
Roles:\allowbreak{} user. Model role:\allowbreak{} panel solver (Qwen2.5-\allowbreak{}7B /\allowbreak{} Llama-\allowbreak{}3.1-\allowbreak{}8B). Parameters:\allowbreak{} \{\allowbreak{}"max\_\allowbreak{}nodes":\allowbreak{} 20,\allowbreak{} "max\_\allowbreak{}tokens":\allowbreak{} 4096,\allowbreak{} "temperature":\allowbreak{} 0.0,\allowbreak{} "top\_\allowbreak{}p":\allowbreak{} 1.0\}\allowbreak{}.
Append JSON source units \{\allowbreak{}id,\allowbreak{}text\}\allowbreak{}; units preserve original line offsets,\allowbreak{} excluding Final Answer-\allowbreak{}only lines. max\_\allowbreak{}nodes=\allowbreak{}min(max\_\allowbreak{}graph\_\allowbreak{}nodes,\allowbreak{}max\_\allowbreak{}support\_\allowbreak{}nodes)=\allowbreak{}20 from the frozen PipelineConfig.
\begin{DDRGPrompt}
\footnotesize\raggedright\sloppy\ttfamily
Extract a reasoning dependency graph from the supplied source units.
\par
Do not solve the question again. Do not correct mistakes in the source.
\par
For each node,\allowbreak{} set source\_\allowbreak{}step to the exact integer id of the selected source unit.
\par
The parser copies that entire source unit verbatim,\allowbreak{} preserving errors and exact offsets.
\par
Do not generate a quote or retype calculations; return only source references and dependencies.
\par
Do not add facts,\allowbreak{} new calculations,\allowbreak{} a new final answer,\allowbreak{} or another candidate.
\par
Use node IDs N1,\allowbreak{} N2,\allowbreak{} ...; parents must refer only to nodes defined earlier.
\par
Every node MUST include parents; use an empty parents list for a premise. Keep at most \{\allowbreak{}max\_\allowbreak{}nodes\}\allowbreak{} nodes.
\par
answer\_\allowbreak{}support must contain the node IDs supporting the source's conclusion.
\par
Return JSON only,\allowbreak{} with exactly these fields:\allowbreak{}
\par
\{\allowbreak{}"nodes":\allowbreak{}[\allowbreak{}\{\allowbreak{}"id":\allowbreak{}"N1",\allowbreak{}"source\_\allowbreak{}step":\allowbreak{}1,\allowbreak{}"parents":\allowbreak{}[\allowbreak{}]\allowbreak{}\}\allowbreak{}]\allowbreak{},\allowbreak{}
\par
 "answer\_\allowbreak{}support":\allowbreak{}[\allowbreak{}"N1"]\allowbreak{}\}\allowbreak{}
\par
Source units:\allowbreak{}
\end{DDRGPrompt}

\paragraph{DDRG alignment}
Roles:\allowbreak{} user. Model role:\allowbreak{} panel solver (Qwen2.5-\allowbreak{}7B /\allowbreak{} Llama-\allowbreak{}3.1-\allowbreak{}8B). Parameters:\allowbreak{} \{\allowbreak{}"max\_\allowbreak{}tokens":\allowbreak{} 4096,\allowbreak{} "temperature":\allowbreak{} 0.0,\allowbreak{} "top\_\allowbreak{}p":\allowbreak{} 1.0\}\allowbreak{}.
graph views serialized by core.make\_\allowbreak{}support\_\allowbreak{}graph\_\allowbreak{}samples and joined with two newlines,\allowbreak{} three hyphens,\allowbreak{} two newlines; source-\allowbreak{}defined wrapper in verifier\_\allowbreak{}first.py.
\begin{DDRGPrompt}
\footnotesize\raggedright\sloppy\ttfamily
Align similar steps in the sampled reasoning graphs.
\par

\par
Do not solve the problem. Do not choose the correct answer.
\par
Only match nodes that talk about the same local fact or step.
\par
Prefer nodes close to ANS. Keep only useful alignments.
\par

\par
Return valid JSON only:\allowbreak{}
\par
\{\allowbreak{}
\par
  "alignments":\allowbreak{} [\allowbreak{}
\par
    \{\allowbreak{}
\par
      "id":\allowbreak{} "A1",\allowbreak{}
\par
      "topic":\allowbreak{} "what these nodes are about",\allowbreak{}
\par
      "claims":\allowbreak{} [\allowbreak{}
\par
        \{\allowbreak{}"graph":\allowbreak{} 1,\allowbreak{} "node":\allowbreak{} "N2",\allowbreak{} "answer":\allowbreak{} "A",\allowbreak{} "claim":\allowbreak{} "short claim"\}\allowbreak{},\allowbreak{}
\par
        \{\allowbreak{}"graph":\allowbreak{} 2,\allowbreak{} "node":\allowbreak{} "N3",\allowbreak{} "answer":\allowbreak{} "B",\allowbreak{} "claim":\allowbreak{} "short claim"\}\allowbreak{}
\par
      ]\allowbreak{}
\par
    \}\allowbreak{}
\par
  ]\allowbreak{}
\par
\}\allowbreak{}
\par

\par

\par
Question:\allowbreak{}
\par
\{\allowbreak{}question\}\allowbreak{}
\par

\par
Parsed support graphs:\allowbreak{}
\par

\par
\{\allowbreak{}graph\_\allowbreak{}views\_\allowbreak{}joined\_\allowbreak{}by\_\allowbreak{}separator\}\allowbreak{}
\end{DDRGPrompt}

\paragraph{DDRG localization}
Roles:\allowbreak{} user. Model role:\allowbreak{} panel solver (Qwen2.5-\allowbreak{}7B /\allowbreak{} Llama-\allowbreak{}3.1-\allowbreak{}8B). Parameters:\allowbreak{} \{\allowbreak{}"max\_\allowbreak{}tokens":\allowbreak{} 4096,\allowbreak{} "temperature":\allowbreak{} 0.0,\allowbreak{} "top\_\allowbreak{}p":\allowbreak{} 1.0\}\allowbreak{}.
The completed Qwen cache has separate alignment\_\allowbreak{}calls and localization\_\allowbreak{}calls; no claim that the unused joint path was evaluated.
\begin{DDRGPrompt}
\footnotesize\raggedright\sloppy\ttfamily
Localize answer-\allowbreak{}sensitive conflicts in aligned answer-\allowbreak{}support graphs.
\par

\par
Return one valid JSON object only. Do not use markdown or code fences.
\par
Use this exact top-\allowbreak{}level schema:\allowbreak{}
\par
\{\allowbreak{}
\par
  "conflicts":\allowbreak{} [\allowbreak{}
\par
    \{\allowbreak{}
\par
      "id":\allowbreak{} "C1",\allowbreak{}
\par
      "alignment":\allowbreak{} "alignment id if available",\allowbreak{}
\par
      "error\_\allowbreak{}type":\allowbreak{} "numeric,\allowbreak{} logical,\allowbreak{} entity,\allowbreak{} relation,\allowbreak{} option\_\allowbreak{}mapping,\allowbreak{} unsupported\_\allowbreak{}jump,\allowbreak{} no\_\allowbreak{}error,\allowbreak{} or unknown",\allowbreak{}
\par
      "left":\allowbreak{} \{\allowbreak{}"graph":\allowbreak{} 1,\allowbreak{} "node":\allowbreak{} "N1",\allowbreak{} "answer":\allowbreak{} "A",\allowbreak{} "claim":\allowbreak{} "short disputed claim"\}\allowbreak{},\allowbreak{}
\par
      "right":\allowbreak{} \{\allowbreak{}"graph":\allowbreak{} 2,\allowbreak{} "node":\allowbreak{} "N1",\allowbreak{} "answer":\allowbreak{} "B",\allowbreak{} "claim":\allowbreak{} "short disputed claim"\}\allowbreak{},\allowbreak{}
\par
      "probe\_\allowbreak{}question":\allowbreak{} "a local question answerable from the original problem"
\par
    \}\allowbreak{}
\par
  ]\allowbreak{}
\par
\}\allowbreak{}
\par

\par

\par
Question:\allowbreak{}
\par
\{\allowbreak{}question\}\allowbreak{}
\par

\par
Parsed answer-\allowbreak{}support graphs:\allowbreak{}
\par

\par
\{\allowbreak{}graph\_\allowbreak{}views\_\allowbreak{}joined\_\allowbreak{}by\_\allowbreak{}separator\}\allowbreak{}
\par

\par
Aligned meta-\allowbreak{}graph:\allowbreak{}
\par
\{\allowbreak{}meta\_\allowbreak{}graph\_\allowbreak{}json\_\allowbreak{}indent2\}\allowbreak{}
\end{DDRGPrompt}

\paragraph{DDRG official local verification}
Roles:\allowbreak{} user. Model role:\allowbreak{} official Qwen3-\allowbreak{}30B-\allowbreak{}A3B-\allowbreak{}Instruct-\allowbreak{}2507,\allowbreak{} registered official model identity. Parameters:\allowbreak{} \{\allowbreak{}"max\_\allowbreak{}tokens":\allowbreak{} 4096,\allowbreak{} "temperature":\allowbreak{} 0.6,\allowbreak{} "top\_\allowbreak{}p":\allowbreak{} 0.95\}\allowbreak{}.
Generated from actual frozen serializer using symbolic inputs. Candidate answer labels,\allowbreak{} gold and ANS nodes removed from payload; source claims may naturally mention option names. Single user message; native template precloses thinking (enable\_\allowbreak{}thinking=\allowbreak{}False). Five class sequences scored under terminal verdict prefix; generated/\allowbreak{}scored agreement required.
\begin{DDRGPrompt}
\footnotesize\raggedright\sloppy\ttfamily
Verify a localized reasoning disagreement between two candidate reasoning paths using ONLY the original problem statement.
\par

\par
[\allowbreak{}Problem Statement]\allowbreak{}
\par
\{\allowbreak{}question\}\allowbreak{}
\par

\par
[\allowbreak{}Local Probe]\allowbreak{}
\par
\{\allowbreak{}probe\_\allowbreak{}question\}\allowbreak{}
\par

\par
[\allowbreak{}LEFT Candidate Evidence]\allowbreak{}
\par
Prerequisites:\allowbreak{}
\par
-\allowbreak{} \{\allowbreak{}left\_\allowbreak{}prerequisites\}\allowbreak{}
\par
Disputed Claim:\allowbreak{}
\par
\{\allowbreak{}left\_\allowbreak{}disputed\_\allowbreak{}claim\}\allowbreak{}
\par

\par
[\allowbreak{}RIGHT Candidate Evidence]\allowbreak{}
\par
Prerequisites:\allowbreak{}
\par
-\allowbreak{} \{\allowbreak{}right\_\allowbreak{}prerequisites\}\allowbreak{}
\par
Disputed Claim:\allowbreak{}
\par
\{\allowbreak{}right\_\allowbreak{}disputed\_\allowbreak{}claim\}\allowbreak{}
\par

\par
[\allowbreak{}Instructions]\allowbreak{}
\par
1. Read the Problem Statement and Local Probe carefully.
\par
2. In 'independent\_\allowbreak{}derivation',\allowbreak{} step-\allowbreak{}by-\allowbreak{}step calculate or deduce the true local fact requested by the Local Probe directly from the Problem Statement. Do NOT rely on either candidate's claims.
\par
3. Compare the LEFT Disputed Claim and prerequisites against your independent derivation:\allowbreak{}
\par
   -\allowbreak{} Is LEFT logically and mathematically sound according to the problem?
\par
4. Compare the RIGHT Disputed Claim and prerequisites against your independent derivation:\allowbreak{}
\par
   -\allowbreak{} Is RIGHT logically and mathematically sound according to the problem?
\par
Assessment discipline (applies to both sides):\allowbreak{}
\par
   -\allowbreak{} Assess only the exact Disputed Claim shown for that side. Quote its relevant words before evaluating them; do not replace it with a different node or with an entire candidate answer.
\par
   -\allowbreak{} Candidate prerequisites are untrusted context,\allowbreak{} not axioms. Check any dependency you use against the Problem Statement.
\par
   -\allowbreak{} Relevance or completeness alone is not evidence of an error. A true intermediate fact is not false merely because it does not finish the Local Probe.
\par
   -\allowbreak{} Different quantities,\allowbreak{} entities,\allowbreak{} units,\allowbreak{} time spans,\allowbreak{} conditional scopes,\allowbreak{} or reasoning steps do not become contradictory merely because they differ. Keep scope explicit in each assessment.
\par
   -\allowbreak{} A decisive LEFT\_\allowbreak{}CORRECT or RIGHT\_\allowbreak{}CORRECT verdict requires a concrete error or hallucination in the rejected side,\allowbreak{} supported by your independent derivation. Greater detail or a more direct answer is not sufficient.
\par
   -\allowbreak{} Do not invent an error to force a decisive verdict. EQUIVALENT requires the same underlying fact. UNKNOWN covers insufficient evidence or non-\allowbreak{}comparable claims; never force a winner between two supported but different facts.
\par
5. Select exactly one verdict from the following 5 options:\allowbreak{}
\par
   -\allowbreak{} "LEFT\_\allowbreak{}CORRECT":\allowbreak{} LEFT is sound and answers the probe correctly,\allowbreak{} while RIGHT contains an error or hallucination.
\par
   -\allowbreak{} "RIGHT\_\allowbreak{}CORRECT":\allowbreak{} RIGHT is sound and answers the probe correctly,\allowbreak{} while LEFT contains an error or hallucination.
\par
   -\allowbreak{} "EQUIVALENT":\allowbreak{} Both sides express the same underlying semantic or mathematical fact (even if wording,\allowbreak{} phrasing,\allowbreak{} or verbosity differs).
\par
   -\allowbreak{} "BOTH\_\allowbreak{}WRONG":\allowbreak{} Both sides make mathematical or logical errors relative to the problem.
\par
   -\allowbreak{} "UNKNOWN":\allowbreak{} Abstain for insufficient evidence or non-\allowbreak{}comparable claims. If a required fact cannot be established from the question,\allowbreak{} state INSUFFICIENT\_\allowbreak{}EVIDENCE in the derivation/\allowbreak{}evidence. If the sides concern different quantities,\allowbreak{} entities,\allowbreak{} scopes,\allowbreak{} conditions or predicates and do not resolve the same local fact,\allowbreak{} state INCOMPARABLE and explain the mismatch. Neither reason permits repair. Teacher disagreement and invalid output are technical/\allowbreak{}supervision statuses,\allowbreak{} not semantic UNKNOWN labels.
\par

\par
Return JSON only in this format:\allowbreak{}
\par
\{\allowbreak{}
\par
  "independent\_\allowbreak{}derivation":\allowbreak{} "<your independent step-\allowbreak{}by-\allowbreak{}step local deduction>",\allowbreak{}
\par
  "left\_\allowbreak{}assessment":\allowbreak{} "<assessment of LEFT>",\allowbreak{}
\par
  "right\_\allowbreak{}assessment":\allowbreak{} "<assessment of RIGHT>",\allowbreak{}
\par
  "relation":\allowbreak{} "",\allowbreak{}
\par
  "verdict":\allowbreak{} "LEFT\_\allowbreak{}CORRECT" | "RIGHT\_\allowbreak{}CORRECT" | "EQUIVALENT" | "BOTH\_\allowbreak{}WRONG" | "UNKNOWN"
\par
\}\allowbreak{}
\end{DDRGPrompt}

\section{Verifier and Decision Details}
\label{sec:decision_details}

\subsection{Five-verdict Interface and Scoring}
The frozen verdict values are \texttt{LEFT\_CORRECT}, \texttt{RIGHT\_CORRECT}, \texttt{BOTH\_WRONG}, \texttt{EQUIVALENT}, and \texttt{UNKNOWN}. Left/right identify the supplied source claims. \texttt{BOTH\_WRONG} rejects both, \texttt{EQUIVALENT} identifies equivalent claims, and \texttt{UNKNOWN} abstains. These outputs do not certify the final correctness of either candidate answer.

\paragraph{Native five-way scoring and gates.}
Verification uses the model's native message template. The aligned scorer conditions on the prefix immediately before the generated verdict value. Each class continuation includes its closing quotation mark. Its score is the sum of conditional token log probabilities, without length normalization; a softmax over the five summed scores gives normalized conditional class scores. These scores compare the registered verdict alternatives and are not calibrated probabilities of factual correctness. The margin is the largest normalized score minus the second largest. The generated class must equal the highest-scored class. Parsing, claim grounding, normalized score at least 0.5, and margin at least 0 are all required; the zero margin threshold adds no strict separation beyond the argmax check, and self-reported confidence is not the gate. Only left/right correctness can yield decisive correction evidence. The actual verifier messages, derivation and verdict formatting, and scoring convention are reproduced in Appendix~\ref{sec:full_prompts}.

\paragraph{Response schema.}
The frozen full response has five JSON fields: \texttt{independent\_derivation}, \texttt{left\_assessment}, \texttt{right\_assessment}, \texttt{relation}, and \texttt{verdict}. The first three hold string-valued local evidence and assessments; the last holds a registered verdict. The parser requires terminal JSON, rejects duplicate keys, ignores labels inside thinking blocks, and treats truncated or malformed output as nondecisive. \texttt{UNKNOWN} includes insufficient evidence and incomparable claims; neither permits repair. True claims about different quantities, entities, or scopes are not a decisive conflict merely because they differ. The complete response template appears in Appendix~\ref{sec:full_prompts}.

\subsection{Grounding, Eligibility, and Abstention}
Extracted claims must quote source text, and alignment/localization references must resolve to those source nodes. Terminal answers are excluded from verification targets. The representative majority graph is fixed by earliest majority candidate position before its usability is tested. The controller follows inferred parent dependencies back from its answer node. Eligible evidence must refute a nonterminal ancestor on this support path and contain a nonempty line of independent derivation. Any decisive support for the majority suppresses correction. Eligible evidence must agree on a single supported candidate-answer provenance value. The frozen controller tests this agreement without an additional nonempty-answer gate; no decisive probe has empty supported provenance in the reported native conditions.

Refuted claims and dependency descendants are excluded from retained context. Remaining nonterminal source claims and provenance accompany the eligible evidence into a single solver regeneration. The controller uses candidate-answer provenance for eligibility, whereas the verifier input and regeneration evidence omit candidate final-answer fields. Graph abstention, unresolved disagreements, failed verification gates, incompatible alternatives, or absent eligible evidence return SC. A parse failure after regeneration also returns SC. Every parseable regeneration is adopted, including a new answer outside the bank; service retries do not create additional logical attempts.

\section{Supplementary Results}
\label{sec:supplementary}

The supplementary audit records the execution units behind the main results
without introducing additional outcome tables. Across the 5,868 registered
Qwen predictions, the source-bound funnel records 200 pre-regeneration
fallbacks, 173 labeled regeneration attempts, and one unparseable regenerated
response that returns to SC. Graph, probe, verifier, and question counts are
stored separately because they represent different stages of the pipeline.
The complete comparison and invalid-output rules are given in Appendix~\ref{sec:complete_setup}. Hidden-test status and native Llama block identities are retained in the author-side manifests.

\subsection{Inference Allocation Accounting}
\label{sec:cost_details}
The source-bound native Qwen DDRG audit covers 61/61 registered blocks and
5,868/5,868 predictions. Total recorded inference allocation is
239,681.909356 GPU seconds (66.5783 GPU hours), including 202,591.770123 GPU
seconds for the 4,868 labeled questions and 37,090.139233 GPU seconds for
hidden ReClor predictions. The allocation is reported as a reproducibility
quantity for the registered execution; it is not used to frame an efficiency
claim against baselines with different accounting bases.

\subsection{Retrospective Candidate-selection Control}
\label{sec:selection_control}
The frozen majority representative, source graphs, recorded native local
probes, grounding checks, generated/scored agreement, thresholds, and path
eligibility were replayed for all 5,868 questions. At eligible questions, the
control returns the unique candidate answer supported by the same evidence
without requesting regeneration; otherwise it returns SC. The replay matched
all 213 recorded regeneration triggers and used no new model calls or
gold-informed selection. On labeled questions, selection yields 31 net
additional correct answers and 72.42\% Qwen macro accuracy, while the actual
fresh regeneration yields 34 net improvements and 72.74\%. This compact control
shows the distinct roles of evidence routing and the final solver response.

\subsection{Post-hoc Statistical Calculation}
\label{sec:statistical_details}
For benchmark $b$, let $W_b$ and $R_b$ denote wrong-to-right and right-to-wrong transitions, and $N_b$ the complete question count. The primary descriptive gain weights benchmarks equally:
\begin{equation}
\widehat\Delta_{\mathrm{macro}}=\frac{1}{5}\sum_{b=1}^{5}\frac{W_b-R_b}{N_b}.
\label{eq:macro_gain}
\end{equation}
The pooled gain instead divides the total net corrections by the total number of questions. Both formulas use accuracy fractions; multiply by 100 to report percentage-point gains.

All 4,868 labeled Qwen questions are joined by exact question ID and frozen bank hash. With $W=63$ and $R=29$, the pooled two-sided exact McNemar calculation is
\paragraph{Paired McNemar calculation.}
\begin{equation}
p=\min\left(1,\;2\sum_{k=0}^{29}
\binom{92}{k}2^{-92}\right)=0.00050909.
\end{equation}
It tests the pooled endpoint, while the primary macro gain weights benchmarks equally. For each bootstrap replicate, questions are resampled with replacement within each of the five benchmarks, preserving paired SC and DDRG correctness. The macro statistic in Equation~\ref{eq:macro_gain} is recomputed from the resampled paired differences. Invalid predictions retain their original denominators. The calculation uses 20,000 replicates and seed 20261008, and the 2.5th and 97.5th percentiles give $[0.1447,1.1504]$ percentage points.

These are retrospective diagnostics conditional on one frozen bank and decoding seed. They assume independent question sampling within benchmarks, do not measure generation-seed stability or uncertainty from method development, and do not establish graph causality. Hidden-label ReClor questions enter neither scoring nor uncertainty. Per-cohort tests are exploratory. Their full source records remain in the author-side audit; no independently replicated multi-seed inference is claimed.

\section{Case Evidence and Source Excerpts}
\label{sec:full_cases}
The following cases were selected to explain successful and harmful correction, not to estimate their prevalence. Table~\ref{tab:case_endpoints} lists the recorded final answers. The labeled correct-retention case is described in Section~\ref{subsec:case_study}. Quoted claims and recorded answers are preserved from the supplied source-bound case material. Figure~\ref{fig:case_gsm8k} abbreviates the claims and compresses their answer-support paths; Figure~\ref{fig:case_failure} is an author-rendered logical explanation. Neither is a claim to reproduce the full raw graph topology.

\subsection{Successful Correction: GSM8K Test 147}
\label{sec:positive_cases}
\paragraph{Original question.}
\begin{DDRGQuote}[blue]
Debra is monitoring a beehive to see how many bees come and go in a day. She sees 30 bees leave the hive in the first 6 hours, and then she sees 1/2 that many bees return in the next 6 hours. She sees two times as many bees as she saw first leave the hive fly from the hive and leave in the next 6 hours. Then every bee that left before that hadn't already returned returns to the hive in the next 6 hours. How many bees did Debra see return to the hive in the last 6 hours of the day?
\end{DDRGQuote}
The recorded bank is $[90,15,75]$, the SC answer is 90, and the gold answer is 75. The issue is the inclusion of bees that have \emph{already returned}, not an incorrect addition operation.

\paragraph{Competing source claims.}
The selected trace's graph 1, node N4 states:
\begin{DDRGQuote}[coral]
So, $30+60=90$ bees return in the last 6 hours.
\end{DDRGQuote}
The competing trace's graph 3, node N9 states:
\begin{DDRGQuote}[teal]
All bees that left earlier (which is the initial 30 minus those that have returned, i.e., $30-15=15$) plus the new bees that left (60), which gives us $15+60=75$ bees that have not yet returned. Therefore, 75 bees return in the last 6 hours.
\end{DDRGQuote}
Only mathematical typesetting and list punctuation are normalized in these excerpts. The normalized conditional score for the recorded \texttt{RIGHT\_CORRECT} verdict is 0.706035, not an independently calibrated probability of truth. Its independent derivation subtracts the 15 earlier returns from the 90 total departures. The controller accepts the refutation of the SC support path, and the saved regeneration ends with \texttt{Final Answer: 75}.

\begin{table}[t]
\begin{nolinenumbers}
\centering
\begingroup\normalsize\setlength{\tabcolsep}{5pt}
\begin{tabular}{@{}lrr@{}}
\toprule
Method & GSM8K 147 & FOLIO 79\\
\midrule
Direct & 90 & A\\
CoT & 90 & B\\
SC@3 & 90 & B\\
GoT (QA) & 45 & B\\
GoV (QA) & 90 & B\\
\rowcolor{DDRGTealLight}
DDRG & 75 & C\\
\midrule
Gold & 75 & B\\
\bottomrule
\end{tabular}\endgroup
\caption{Recorded final answers for the successful and harmful cases. The table shows endpoint comparisons, not full intermediate traces for every baseline.}
\label{tab:case_endpoints}
\end{nolinenumbers}
\end{table}

\subsection{Harmful Correction: FOLIO Validation 79}
\paragraph{Question and answers.}
The premises state that no bakery is spicy, all cupcakes are bakeries, all hotpots are spicy, and all products of Baked by Melissa are cupcakes. They also state: if dried Thai chilies are spicy or a bakery, then they are a hotpot or are spicy. The conclusion is that dried Thai chilies are a product of Baked by Melissa; the choices are A (True), B (False), and C (Uncertain). The bank is $[B,C,B]$ and gold is B.

\Needspace{11\baselineskip}
\paragraph{Recorded local comparison.}
The SC representative's graph 1, node N16 states:
\begin{DDRGQuote}[blue]
Since dried Thai chilies can be either spicy or a hotpot (the latter being consistent with not being a bakery), but the key point is that they cannot be both a bakery and spicy at the same time, we need to determine if they can be a product of Baked by Melissa:
\end{DDRGQuote}
Graph 2, node N12 instead states:
\begin{DDRGQuote}[orange]
The premises do not provide direct information linking dried Thai chilies to being a product of Baked by Melissa. They only establish that if dried Thai chilies were spicy or a bakery, they would be a hotpot or spicy, but we have no information to confirm or deny that dried Thai chilies fit into these categories.
\end{DDRGQuote}
The verifier accepts the latter direction with normalized conditional score 0.651367. The controller triggers a regeneration, which ends with \texttt{Final Answer: Uncertain}; the normalized answer is C.

\paragraph{Why the correction fails.}
Under the stated premises, being a product implies being a cupcake, hence a bakery and not spicy. But for dried Thai chilies, being a bakery implies being a hotpot or spicy; either alternative implies spicy. This contradiction rules out the product conclusion. The saved verifier derivation misses that consequence and treats missing direct information as uncertainty. Thus the source match and numerical gate pass while the semantic judgment fails. Figure~\ref{fig:case_failure} shows this explanatory chain; the full system result remains a harmful B-to-C change.

\end{document}